%% file: main.tex
\documentclass{article}
\usepackage{amssymb}
\usepackage{amsmath}
\usepackage{amsthm}
\usepackage{graphicx}
\usepackage[font=small]{caption}
\usepackage{subcaption}
\usepackage{bm}
\usepackage{comment}
\usepackage{wrapfig}
\usepackage{algorithm}
\usepackage{algorithmic}
\usepackage{booktabs}
\usepackage{enumitem}

\newtheorem{theorem}{Theorem}

\newcommand{\Var}{\mathrm{Var}}
\newcommand{\Cov}{\mathrm{Cov}}

\usepackage[dvipsnames]{xcolor}

\newcommand\muhatmc[0]{$\hat{\mu}_{\mathrm{MC}}$}
\newcommand\muhatbetahat[0]{$\hat{\mu}_\mathrm{CV}$}

\newcommand{\npaired}{n}
\newcommand{\nsurrogate}{k}
\newcommand{\scenario}{X}
\newcommand{\scenariospace}{\mathcal{X}}
\newcommand{\scenariodist}{P_X}
\newcommand{\metric}{F}
\newcommand{\metricspace}{\mathbb{R}}
\newcommand{\metricdist}{P_{F|X}}
\newcommand{\surrogate}{\bm{G}}
\newcommand{\surrogatedim}{d}
\newcommand{\surrogatespace}{\mathbb{R}^\surrogatedim}
\newcommand{\surrogatedist}{P_{\bm{G}|X}}
\newcommand{\surrogatedata}{\mathcal{D}_\mathrm{surrogate}}
\newcommand{\paireddata}{\mathcal{D}_\mathrm{paired}}
\newcommand{\estdata}{\mathcal{D}_\mathrm{est}}
\newcommand{\nest}{n_\mathrm{est}}
\newcommand{\fitdata}{\mathcal{D}_\mathrm{fit}}
\newcommand{\nfit}{n_\mathrm{fit}}
\newcommand{\coeff}{\bm{\beta}}
\newcommand{\mcf}{\hat{f}}
\newcommand{\feat}{\phi}

\usepackage[preprint]{corl_2025} 

\usepackage{titlesec}
\titlespacing{\section}{0mm}{0mm}{0mm}
\titlespacing{\subsection}{0mm}{0mm}{0mm}
\title{Sim2Val: Leveraging Correlation Across Test Platforms for Variance-Reduced Metric Estimation}

\author{
\textbf{Rachel Luo$^{1}$, Heng Yang$^{1,2}$, Michael Watson$^{1}$, Apoorva Sharma$^{1}$, Sushant Veer$^{1}$,} \\
\textbf{Edward Schmerling$^{1}$, Marco Pavone$^{1,3}$} \\
$^1$NVIDIA \quad $^2$Harvard University \quad $^3$Stanford University \\
\texttt{\{raluo, hengy, mwatson, apoorvas, sveer, eschmerling, mpavone\}@nvidia.com}
}

\begin{document}
\def\baselinestretch{0.97}\selectfont
\maketitle


\begin{abstract}
    Learning-based robotic systems demand rigorous validation to assure reliable performance, but extensive real‐world testing is often prohibitively expensive, and if conducted may still yield insufficient data for high-confidence guarantees. In this work we introduce Sim2Val, a general estimation framework that leverages \textit{paired} data across test platforms, e.g., paired simulation and real‐world observations, to achieve better estimates of real-world metrics via the method of control variates. By incorporating cheap and abundant auxiliary measurements (for example, simulator outputs) as control variates for costly real‐world samples, our method provably reduces the variance of Monte Carlo estimates and thus requires significantly fewer real‐world samples to attain a specified confidence bound on the mean performance. We provide theoretical analysis characterizing the variance and sample-efficiency improvement, and demonstrate empirically in autonomous driving and quadruped robotics settings that our approach achieves high‐probability bounds with markedly improved sample efficiency. Our technique can lower the real‐world testing burden for validating the performance of the stack, thereby enabling more efficient and cost‐effective experimental evaluation of robotic systems. 
\end{abstract}



\section{Introduction}
\vspace{-1mm}

Rigorous evaluation of autonomous systems is essential for both research and real-world deployment. However, conducting comprehensive real-world tests is often prohibitively expensive and thus, acquiring a sufficient number of real-world samples to ensure high-confidence estimates of performance metrics (as is necessary for validation) is often infeasible~\citep{Kalra2016}.
To mitigate these challenges, system designers frequently rely on alternative test platforms --- such as simulators or historical data from previous versions of a robot's policy --- to obtain cost-effective, scalable data. However, these auxiliary domains may not accurately reflect performance of the current system in the target deployment environment, and treating outputs from these sources as equivalent to real data can therefore lead to misleading performance estimates. Instead, our goal in this work is to leverage the abundant auxiliary-domain data to more accurately estimate the real-world performance metrics of interest. 

We introduce a general framework, Sim2Val, that leverages correlation across heterogeneous test platforms to improve metric estimation in a target domain. For instance, we may wish to estimate a safety-related real-world autonomous driving metric for validation purposes, but want to leverage our extensive simulation data to do so. Our key insight is to exploit \textit{paired} observations --- instances for which we collect simultaneous measurements on both an auxiliary platform (e.g. a simulator or offline policy logs) and the target platform (e.g. real-world tests). From these paired samples, we can then learn a \textit{metric correlator function} that maps surrogate metrics to real-world metrics, thereby improving the correlation between the two. 
By incorporating the auxiliary outputs as a control variate in a Monte Carlo estimator, we provably reduce variance whenever the two platforms exhibit nontrivial correlation, yielding unbiased performance estimates with substantially fewer target‐domain (e.g., real-world) samples. Notably, this approach works well for auxiliary data from a simulator or from offline logs of an earlier policy: in either case, a small set of paired examples suffices to ``transfer'' cheap measurements into tighter confidence bounds on the metric of interest. We empirically demonstrate our metric estimation framework in autonomous driving and quadruped robotics settings, showing {
\parfillskip=0pt
\parskip=0pt
\par}
\begin{figure}
    \centering
    \includegraphics[width=\linewidth]{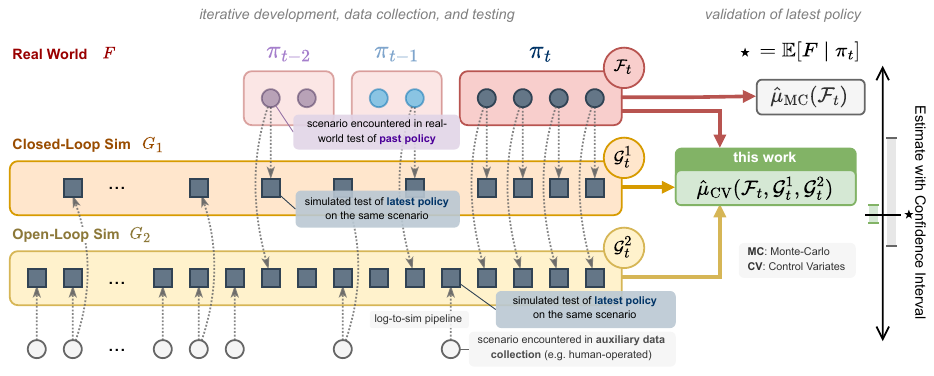}
    \caption{When validating an autonomous system, we can measure performance across multiple test platforms, but fundamentally seek to measure performance in the real-world. While real-world measurements (circles) are expensive, log-to-sim pipelines enable measuring simulated metrics (squares) for the latest policy on scenarios encountered in past tests and data-collection campaigns. In this work, we propose a control variates approach which leverages these additional simulated tests to estimate the expected real-world performance. This method achieves tighter confidence intervals than using real-data alone so long as the simulated metrics are predictive of real-world performance.}
    \label{fig:anchor}
    \vspace{-5mm}
\end{figure}

\begin{wrapfigure}{r}{0.3\textwidth}
  \vspace{-2.5mm}
  \begin{center}
    \includegraphics[width=0.28\textwidth]{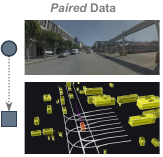}
  \end{center}
  \vspace{-3mm}
  \caption{Paired real world and simulation data.}
  \vspace{-5mm}
\end{wrapfigure}
\noindent that our approach achieves high-probability bounds with markedly improved sample efficiency. 

Thus, our contributions are four-fold: (1) a method for leveraging \textit{paired} data to reduce the variance of Monte Carlo estimates of real-world performance metrics; (2) the adaptation of classical control-variates techniques for robotics applications, enhanced by a metric correlator function to improve performance when the baseline correlation is low; (3) theoretical bounds on variance reduction and sample‐complexity when using a correlated auxiliary signal; and (4) experimental validation of our framework in autonomous driving and quadruped robotics settings.

By leveraging correlation across different test platforms, our approach enables more scalable and stronger validation of robotic systems.


\vspace{2mm}
\section{Background and Related Work}
\label{sec:background}
\vspace{-1mm}

This paper considers the problem of validating an autonomous system in a data-efficient way. In particular, our focus is on estimating the expected value of a metric of interest on the distribution of environmental conditions expected at deployment time, potentially conditioned on a particular subdomain of interest.\footnote{We stress that this form of statistical validation is only one element of full system validation; measuring the tail metrics and targeting known failure modes are also critical, but not the focus of this work.}
Testing autonomous systems in the real world can be expensive; real-world mistakes can be costly, requiring tests to be carefully supervised. Given the costs of real-world testing, one key application of our work is how to use surrogate metrics obtained through log-based simulation to improve our real-world sample efficiency in estimating performance. 

\vspace{-2mm}

\paragraph{Log-Replay Simulation}
Physics-based simulators have provided an accurate test-platform for some robotics tasks, but closing the sim-to-real gap with purely synthetic simulation is challenging, particularly for open-world robotics applications with high-dimensional input spaces.
To address this, an increasingly common paradigm is the concept of ``log-replay simulation,'' which decouples real-world data collection (i.e. obtaining logs) from system evaluation (replaying those logs). Log-replay simulation can be as simple as open-loop evaluation of a policy on scenarios sampled in human-operated data collection \citep{caesar2020nuscenes, karnchanachari2024towards, open_x_embodiment_rt_x_2023}, or as complex as closed-loop simulation with reactive agents \citep{gulino2023waymax, karnchanachari2024towards} and re-simulated sensor data \citep{urtasan2025simulator, wu2025alpamayo} seeded from sensor logs of real-world scenarios. 
Notably, log-replay based simulation enables \textit{paired} measurements between real-world and simulated domains. In this work, we leverage the availability of such paired data to incorporate surrogate-only measurements at scale into statistical estimators of real-world metrics.

\paragraph{Offline (Batch) Reinforcement Learning \& Off-Policy Evaluation}
Motivated by the costs of on-policy data collection, the field of \textit{offline} (or \textit{batch}) reinforcement learning aims to study policies with limited ability to test in closed-loop \citep{levine2020offline}. Most relevant to this work is the topic of \textit{off-policy evaluation}, where the goal is to estimate the performance of the current policy $\pi_t$ using data collected under a different policy. Methods for off-policy evaluation typically leverage importance sampling to re-weight samples in the dataset to account for the difference between $\pi_t$ and the data-collection policy \citep{precup2000eligibility}, or rely on using observed data fitting a model of the environment to obtain surrogate measurements of the performance of $\pi_t$. Of particular relevance to our work is the concept of the \textit{doubly robust estimator} \citep{jiang2016doubly,thomas2016data}, which proposes augmenting the importance sampling approach by leveraging an approximate Q function as a control variate to reduce variance. In this work, we consider the system under test and the behavior policy to be a black box, so we do not use action probabilities for importance sampling, but we use surrogate metrics from a test platform to serve the role of a control variate.

\vspace{-2mm}

\paragraph{Monte-Carlo Methods for Variance Reduction}
Advanced Monte-Carlo methods,
such as importance sampling, stratified sampling, and control variates, provide practitioners with tools to leverage domain knowledge to obtain tighter confidence intervals than na\"ive Monte-Carlo \cite{owen2013monte}. 
In this work, we use the technique of control variates, which reduces variance by leveraging an auxiliary signal that is correlated with the metric of interest, but whose expectation is known. Prediction-powered inference \cite{angelopoulos2023prediction, angelopoulos2023ppi++} applies similar ideas to improve the estimation of statistical quantities by leveraging synthetic labels as control variates, and has been applied to reduce human labeling costs for LLM evaluation \cite{zhou2025accelerating, boyeau2024autoeval}. In this work, we adapt these ideas for autonomous system validation where results on surrogate test platforms serve the role of the control variate or synthetic labels, and highlight how learning a mapping from a surrogate platform metric to the target metric can further tighten confidence bounds.


\vspace{2mm}
\section{Sim2Val Method}
\label{sec:method}
\vspace{-1mm}


\subsection{Problem Setup}\label{subsec:problem-setup}
Let $\scenario \in \scenariospace$ denote a random variable with probability distribution $\scenariodist$ over some environmental conditions. Let $\metric \in \metricspace$ be a random variable corresponding to the value of a metric of interest, with conditional distribution $\scenariodist$.
For example, $\scenario$ may represent the environmental conditions for a driving scenario (road geometry, location of road agents, road agent intents), and $\metric$ could represent the minimum time-to-collision experienced by an AV planner when rolled out in the scenario. 
Our goal is to estimate the expected value of $\metric$ over the joint distribution of $(\scenario, \metric)$,
\begin{align}
\mu \equiv \underset{\scenario \sim \scenariodist}{\mathbb{E}}\left[\underset{\metric \sim \metricdist}{\mathbb{E}}[\metric \mid \scenario]\right].
\end{align}
We assume that we have a measurement of $F$ on $\npaired$ samples from $P_X$, $\{(\scenario_i, \metric_i)\}_{i=1}^{\npaired}$. A na\"ive Monte Carlo estimator of $\mu$ corresponds to the empirical mean,
\begin{align}
\label{eq:mc_estimator}
    \hat{\mu}_\mathrm{MC} &:= \frac1n\sum_{i=1}^\npaired \metric_i, & \Var[\hat{\mu}_\mathrm{MC}] &= \frac1n \Var[ \metric ] \approx \frac{1}{n^2} \sum_{i=1}^N ( F_i - \hat{\mu}_\mathrm{MC} )^2.
\end{align}

Cheaper surrogate test platforms, e.g. log-replay simulation, allow us to measure surrogate metrics $\surrogate \in \surrogatespace$ conditioned on a scenario $\scenario$. These surrogate measurements may be stochastic themselves, so we model them as a random vector with conditional distribution $\surrogatedist$. Importantly, we assume that we can sample $\surrogate$ both for the $\npaired$ scenarios for which we have a measurement of $\metric$, and for additional logged scenarios, e.g., those sampled during tests of other systems or in human-operated data collection. 
Ultimately, we assume that we have $\npaired$ paired samples $\paireddata = \{(X_i, \metric_i, \surrogate_i)\}_{i=1}^\npaired$, and $\nsurrogate$ surrogate-only samples, $\surrogatedata = \{(\scenario'_j, \surrogate'_j)\}_{j=1}^\nsurrogate$, where 
\begin{align*}
    \scenario_i \sim \scenariodist \quad &\forall\,i=1,\dots,\npaired 
    & \scenario'_j \sim \scenariodist \quad & \forall\,j=1,\dots,\nsurrogate \\
    \metric_i \sim \metricdist(\cdot \mid \scenario_i) \quad &\forall\,i=1,\dots,\npaired \\
    \surrogate_i \sim \surrogatedist(\cdot \mid \scenario_i) \quad &\forall\,i=1,\dots,\npaired
    & \surrogate'_j \sim \surrogatedist(\cdot \mid \scenario'_j) \quad &\forall\,j=1,\dots,\nsurrogate
\end{align*}

The nature of $(X_i, \metric_i, \surrogate_i)$ depends on the experimental setting. For example: (i) $X_i$ may be a specific log, with $\metric_i$ denoting the real-world measurement and $\surrogate_i$ the corresponding simulation replay measurement; (ii) $X_i$ may again be a specific log with $\metric_i$ denoting the real-world measurement, but $\surrogate_i$ may be obtained by running simulation on a scenario $X'_i$ retrieved from a simulation database of scenarios similar to $X_i$; or (iii) $X_i$ may represent a semantic category (e.g., urban night driving), with $\metric_i$ denoting the real-world performance and $\surrogate_i$ the average simulation performance on scenarios from the same category. The key requirement is that $\metric$ and $\surrogate$ exhibit nontrivial correlation.

Our goal is to strategically leverage the surrogate metric observations to achieve a tighter estimate for $\mu$. Critically, we make no a priori assumptions on the relationship between $\metricdist$ and $\surrogatedist$; we treat the surrogate test platforms as a black box. Nevertheless, as we will demonstrate, a surrogate platform where $\surrogate$ is highly correlated with $\metric$ can enable more efficient, lower variance statistical estimators for $\mu$.

\begin{figure}[t!]
    \centering
    \begin{subfigure}[t]{0.69\linewidth}
    \centering
    \includegraphics[width=\textwidth]{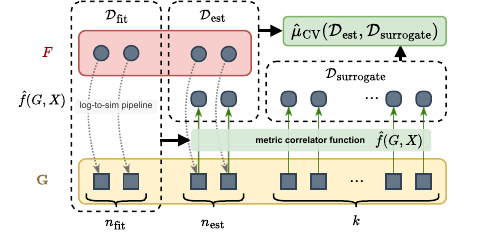}
    \end{subfigure}%
    ~
    \begin{subfigure}[t]{0.3\linewidth}
    \centering
    \includegraphics[width=\textwidth]{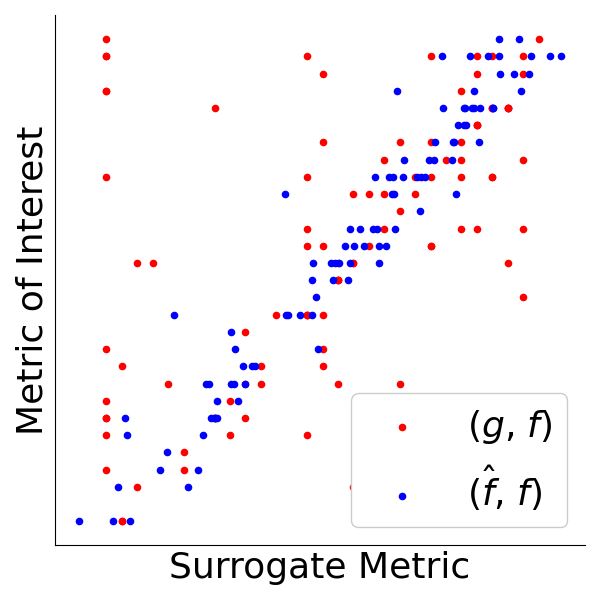}
    \end{subfigure}
    \caption{(Left) Given some paired measurements of our real ($\metric$) and surrogate ($\surrogate$) metrics, our approach uses some of that data $\fitdata$ to learn a \textit{metric correlator function} (MCF, $\mcf$), mapping the surrogate metric and context features to a prediction of the corresponding real-world measurement. Then, we can apply the MCF to the remaining surrogate metric values to construct the paired $\estdata$ and surrogate only $\surrogatedata$ datasets used produce the control-variates estimator of $\mathbb{E}[\metric]$. (Right) Empirical results on the NuPlan dataset showing that fitting the MCF can extract a more predictive signal for $\metric$ from the surrogate metrics $\surrogate$, improving correlation.}
    \vspace{-3mm}
    \label{fig:method}
\end{figure}

\subsection{Control Variate Framework with Surrogate Data}

We adapt the method of control variates to improve our estimation of $\mu$ using the inexpensive surrogate $\surrogate$ as a control variate. Typically, the method of control variates requires perfect knowledge of $\theta := \mathbb{E} [ \mathbb{E}[\surrogate \mid \scenario ] ]$, i.e., the expected value of our surrogate metric; however, in our application the \emph{exact} value of $\theta$ is not accessible. Nevertheless, given that the number of surrogate samples ($\nsurrogate$) is much higher than the number of true metric samples we can collect ($\npaired$), applying this strategy with an estimated value of $\theta$ can produce a tighter variance. Specifically, we define the estimator
\begin{align}
\label{eq:cv-estimator}
    \hat{\mu}_{CV}(\paireddata, \surrogatedata; \coeff) &=  \frac{1}{n} \sum_{i=1}^{n} \left( \metric_i - \coeff^T \surrogate_i \right) +  \frac1{\nsurrogate} \sum_{j=1}^\nsurrogate \coeff^T\surrogate'_j , 
\end{align}
where $\coeff \in \mathbb{R}^\surrogatedim$ is a hyperparameter. This estimator has lower variance than the simple Monte-Carlo estimator when $\surrogate$ is correlated with $\metric$, and the degree of variance reduction depends on the degree of correlation and the number of additional surrogate only samples $\nsurrogate$. 

\vspace{1mm}
\begin{theorem}[Optimal Variance of $\hat{\mu}_\mathrm{CV}$]
    \label{thm:variance-reduction}
    Let $F$ and $\surrogate$ be as defined in Section~\ref{subsec:problem-setup}. Let $\hat{\mu}_{\mathrm{CV}}$ be as defined in \eqref{eq:cv-estimator}.
    Then, if $\coeff$ is chosen as $\coeff_{\rm opt}$, defined as:
    \begin{align}\label{eq:cv_beta_opt}
        \coeff_\mathrm{opt} := \left(\frac{\nsurrogate}{\nsurrogate + \npaired}\right) \Var(\surrogate)^{-1} \Cov(\surrogate, \metric),
    \end{align}
    the variance of $\hat{\mu}_{\mathrm{CV}}$ takes the following form:
    \begin{equation}\label{eq:cv-variance}
        \mathrm{Var}(\hat{\mu}_\mathrm{CV}(\coeff_\mathrm{opt})) = \frac{1}{n} \left( 1 - \frac{\nsurrogate}{\nsurrogate + \npaired} \rho^2(\surrogate,\metric) \right) \Var(\metric),
    \end{equation}
    where $
        \rho^2(\surrogate,\metric) := \| \Var(\surrogate)^{-\frac12} \Cov(\surrogate,\metric) \Var(\metric)^{-\frac12} \|^2_{\rm Fr}$
    and captures how correlated the surrogate metric is to the metric of interest.\footnote{
    Note that in the case of scalar $\surrogate$, $\rho$ corresponds to the Pearson correlation coefficient between $\metric$ and $\surrogate$.}
\end{theorem}

\begin{wrapfigure}{L}{0.48\textwidth}
\vspace{-4mm}
\begin{minipage}{0.47\textwidth}
\begin{algorithm}[H]
\caption{Control Variate Estimator}
\label{alg:cv_method}
\small
\begin{algorithmic}[1]
  \REQUIRE Metric of interest $\metric$, surrogate metrics $\surrogate$, paired samples $\paireddata = \{(X_i, \metric_i, \surrogate_i\}_{i=1}^\npaired$, surrogate-only samples $\surrogatedata = \{(\scenario'_j, \surrogate'_j)\}_{j=1}^\nsurrogate$, desired confidence interval size $\alpha$
  \STATE Estimate the optimal coefficient $\hat\coeff_{\rm opt}$ Eq.~\eqref{eq:cv_beta_opt}, using sample estimates of $\Var(\surrogate)$, $\Cov(\surrogate, \metric)$. \footnote{Note that using the same samples to estimate $\hat\coeff$ and to subsequently compute the mean introduces a very small bias; separate data may be used to estimate $\hat\coeff$ to eliminate this bias if necessary.}
  \STATE Compute the estimate of mean performance $\hat{\mu}_{CV}(\paireddata, \surrogatedata; \hat\coeff_{\rm opt})$, Eq.~\eqref{eq:cv-estimator}.
  \STATE Compute the estimator variance $\widehat{\mathrm{Var}}(\hat{\mu}_\mathrm{CV}(\hat\coeff_\mathrm{opt}))$, Eq.~\eqref{eq:cv-variance}, using sample estimates of $\Var(\metric)$, $\Var(\surrogate)$, $\Cov(\surrogate,\metric)$.
  \STATE Compute the confidence interval $\mathbb{P}\left( \left| \hat{\mu}_{CV}(\paireddata, \surrogatedata; \hat\coeff_{\rm opt}) - \mu \right| \geq \alpha \right)$\\
  $\leq \widehat{\mathrm{Var}}(\hat{\mu}_\mathrm{CV}(\hat\coeff_\mathrm{opt})) / \alpha^2$, Chebyshev or CLT.
\end{algorithmic}
\normalsize
\end{algorithm}
\end{minipage}
\vspace{-4mm}
\end{wrapfigure}

We see that the optimal control variate estimator can do no worse in terms of variance than the simple Monte Carlo estimator, and that as the amount of the surrogate data increases, the variance of the estimator approaches a limit governed by the correlation of $\surrogate$ and $\metric$, $(1-\rho^2)\Var(\metric)$. Algorithm ~\ref{alg:cv_method} summarizes our approach. 

A direct consequence of the lower variance from the control variate estimator is a reduction in the amount of real-world data needed to estimate the same confidence interval around $\mu$. We define a confidence interval (CI) $\gamma:=(\alpha, p)$ by the tuple that includes a measure of the ``size" $\alpha\in[0,\infty)$ of the interval in which we expect the true mean to lie and the probability $p\in[0,1]$ of the interval \emph{not} covering the true mean. A CI $\gamma_1:=(\alpha_1, p_1)$ is considered to be tighter than $\gamma_2:=(\alpha_2, p_2)$ if $\alpha_1 \leq \alpha_2$ while $p_1 \leq p_2$. With this notion of CI tightness, we present the following theorem.

\vspace{1mm}
\begin{theorem}[Sample Efficiency for CI Estimation with Chebyschev Inequality]\label{thm:sample-complexity}
    Let $n_r\in\mathbb{Z}_{>0}$ be the number of real-world samples of $F$ for direct estimation of $\mu$ (as in Eq.~\ref{eq:mc_estimator}). Let $n_p\in\mathbb{Z}_{>0}$ be the number of real-world and surrogate data pairs $(F,\surrogate)$ drawn for estimating $\mu$ using control variates (as in Eq.~\ref{eq:cv-estimator} and~\ref{eq:cv_beta_opt}). 
    Let $k$ be the number of surrogate-only samples. If we use the Chebyschev inequality to estimate $\mu$, then
    \begin{enumerate}[label=(\roman*), leftmargin=3em]
        \item if $n_r=n_p$, the CI from control variates is tighter than the CI from direct estimation;
        \item to obtain a CI of equivalent width and confidence, we require $n_p$ to be at least $n_{\min}$ where
        \begin{equation}\label{eq:cv-data-reduction}
            n_{\min}:= \frac{-(k-n_r) + \sqrt{(k-n_r)^2 + 4n_rk(1-\rho^2)}}{2} \leq n_r.
        \end{equation}
    \end{enumerate}
\end{theorem}
The proofs of Theorem~\ref{thm:variance-reduction} and Theorem~\ref{thm:sample-complexity} are provided in Appendix~\ref{app:proofs}.\footnote{While Theorem~\ref{thm:sample-complexity} is stated using Chebyshev’s inequality, similar conclusions hold for any confidence interval construction whose width scales with the variance. In practice, confidence intervals can be constructed from the control variate samples using any standard approach (such as Chebyshev, Central Limit Theorem, or Hoeffding bounds). CLT-based intervals are typically the tightest, as they exploit asymptotic normality.}

\vspace{1mm}
\subsection{Improving Performance by Learning a Metric Correlator Function}

\begin{wrapfigure}{L}{0.49\textwidth}
\vspace{-8mm}
\begin{minipage}{0.48\textwidth}
\begin{algorithm}[H]
\caption{Improved Control Variate Estimation with a Metric Correlator Function}
\label{alg:cv_mcf}
\small
\begin{algorithmic}[1]
  \REQUIRE Metric of interest $\metric$, surrogate metrics $\surrogate$, paired samples $\paireddata = \{(X_i, \metric_i, \surrogate_i\}_{i=1}^\npaired$, surrogate-only samples $\surrogatedata = \{(\scenario'_j, \surrogate'_j)\}_{j=1}^\nsurrogate$, scenario features $\feat(\scenario)$, desired confidence interval size $\alpha$
  \STATE Split the $\npaired$ paired samples 
  into $\fitdata$ of size $\nfit$ and $\estdata$ of size $\nest$, such that $\npaired = \nfit + \nest$.
  \STATE Train a metric correlator function $\mcf: \bigl(\surrogate,\,\feat(\scenario)\bigr) \;\longrightarrow\;\metric$ using the samples in $\fitdata$.
  \STATE Apply Algorithm~\ref{alg:cv_method}, with $\surrogate := \mcf(\surrogate, \feat(\scenario))$ and $\npaired := \nest$.
\end{algorithmic}
\normalsize
\end{algorithm}
\end{minipage}
\vspace{-1mm}
\end{wrapfigure}
The efficacy of the control variate method described in Algorithm~\ref{alg:cv_method} relies on the correlation of the surrogate metrics $\surrogate$ to the true metric of interest $\metric$ conditioned on a scenario $\scenario$. Although we would hope to ensure that these are strongly correlated when we design our surrogate test platforms (e.g., by designing predictive open-loop metrics \citep{dauner2024navsim}, or improving the fidelity of our simulator \citep{urtasan2025simulator}), there may still be a benefit in considering additional features of $\scenario$. To further improve the correlation, we propose learning a \textit{metric correlator function} (MCF) $\mcf$ to map the surrogate measurements $\surrogate$ as well as features of the scenario $\feat(\scenario)$ to a prediction of the metric of interest $\metric$. As shown in Figure \ref{fig:method}, we partition the paired data $\paireddata$ into $\fitdata$ of size $\nfit$, and $\estdata$ of size $\nest := \npaired - \nfit$, and train the MCF on $\nfit$ with an MSE loss.

We predict the metric of interest $\hat{\metric}$ for all remaining data. Then, we can apply the same control variate estimator algorithm as above, replacing $\surrogate$ with $\mcf(\surrogate, \feat(\scenario))$ (and $\npaired$ with $\nest$). In our experiments, we find that training a metric correlator function in this manner can dramatically increase correlation. However, this comes at the cost of losing some paired data; the samples that are used in training the metric correlator function cannot be used in the computation of the estimate itself. The MCF-enhanced estimator provides a net variance reduction so long as 
\begin{equation}
    \label{eq:mcf_reduction_criterion}
    \rho^2_\mathrm{MCF}/\left(1 + \frac{\nest}{\nsurrogate}\right) > \rho^2 / \left( 1 + \frac{\npaired}{\nsurrogate}\right).
\end{equation}

Note, however, that the MCF can be trained on any data where both the true metric and surrogate features are available; the training procedure is not limited to samples from $\paireddata$. For example, data from related but non-target domains may be leveraged for training if the learned correlations generalize well to the target domain. Importantly, leveraging data from outside $\paireddata$ does not consume any of the paired samples from $\paireddata$ that would otherwise be used in the control variate estimation (i.e., it does not affect $\nest$). Algorithm~\ref{alg:cv_mcf} outlines our improved estimation procedure incorporating the learned MCF.


\vspace{2mm}
\section{Experimental Results}
\label{sec:result}

We evaluate our variance‐reduction framework on three distinct settings: (1) closed‐loop versus open‐loop metrics in the nuPlan autonomous driving dataset, (2) autonomous vehicle (AV) simulator outputs for predicting real‐world driving performance, and (3) simulator outputs for predicting real-world measurements in a quadruped locomotion task. In each case, we compare the sample variance of our control‐variates estimator against the standard Monte Carlo baseline, demonstrating consistent reductions in uncertainty around the estimated mean. Furthermore, by learning an MCF, we observe additional variance reduction, particularly in domains where the raw correlation is low.

\subsection{nuPlan Simulation Experiments}
We first consider the nuPlan dataset~\citep{nuplan}, which provides large‐scale open‐loop and closed-loop simulation metrics (e.g.\ time to collision (TTC), average displacement error (ADE), drivable area compliance, etc.), for various high‐level scenario classes (e.g.\ ``on\_intersection,'' ``near\_multiple\_vehicles''). 
Here, open-loop simulation employs non-reactive agents following a fixed trajectory (and is thus computationally inexpensive), whereas closed‐loop simulation uses fully reactive agents that respond to their environment (and is thus computationally expensive). We demonstrate how our approach can leverage results on the cheap open-loop simulation to improve estimates of performance in closed-loop.

Within each scenario class,
we draw a limited set of $\npaired$ paired samples 
\( \{ ( \scenario_i, \surrogate_i, \metric_i ) \}_{i=1}^\npaired \)
where $\surrogate_i$ and $\metric_i$ denote the open-loop and closed-loop metrics respectively sampled for $\scenario_i$. Furthermore, we draw an additional \(\nsurrogate\) unpaired open-loop observations \(\{(\scenario'_j, \surrogate'_j\}_{j=1}^{k}\).  
We compare three estimators for the population mean \(\mu = \mathbb{E}[f(X)]\):   
\begin{enumerate}
  \item \textbf{Monte Carlo baseline (MC):} \(\hat\mu_{\mathrm{MC}} = \tfrac{1}{n}\sum_{i=1}^n \metric_i\).
  \item \textbf{Control variates (CV):} \(\hat\mu_{CV} = \frac{1}{\npaired}\sum_{i=1}^\npaired (\metric_i - \hat\coeff^T \surrogate_i)  + \frac1\nsurrogate \sum_{j=1}^\nsurrogate \hat\coeff^T\surrogate'_j \), with \(\hat\coeff\) chosen as described in Section~\ref{sec:method}.
  \item \textbf{Control variates with a metric correlator function (CV-MCF):} We replace \(\surrogate\) by \(\mcf(\surrogate, \feat(\scenario))\), where $\nfit$ of the paired samples are used to fit $\mcf$, and the remaining $\nest = \npaired-\nfit$ paired samples are used in the control variate estimator.
\end{enumerate}

\begin{figure} 
    \centering
    \begin{subfigure}{0.44\textwidth}
    \includegraphics[width=\textwidth]{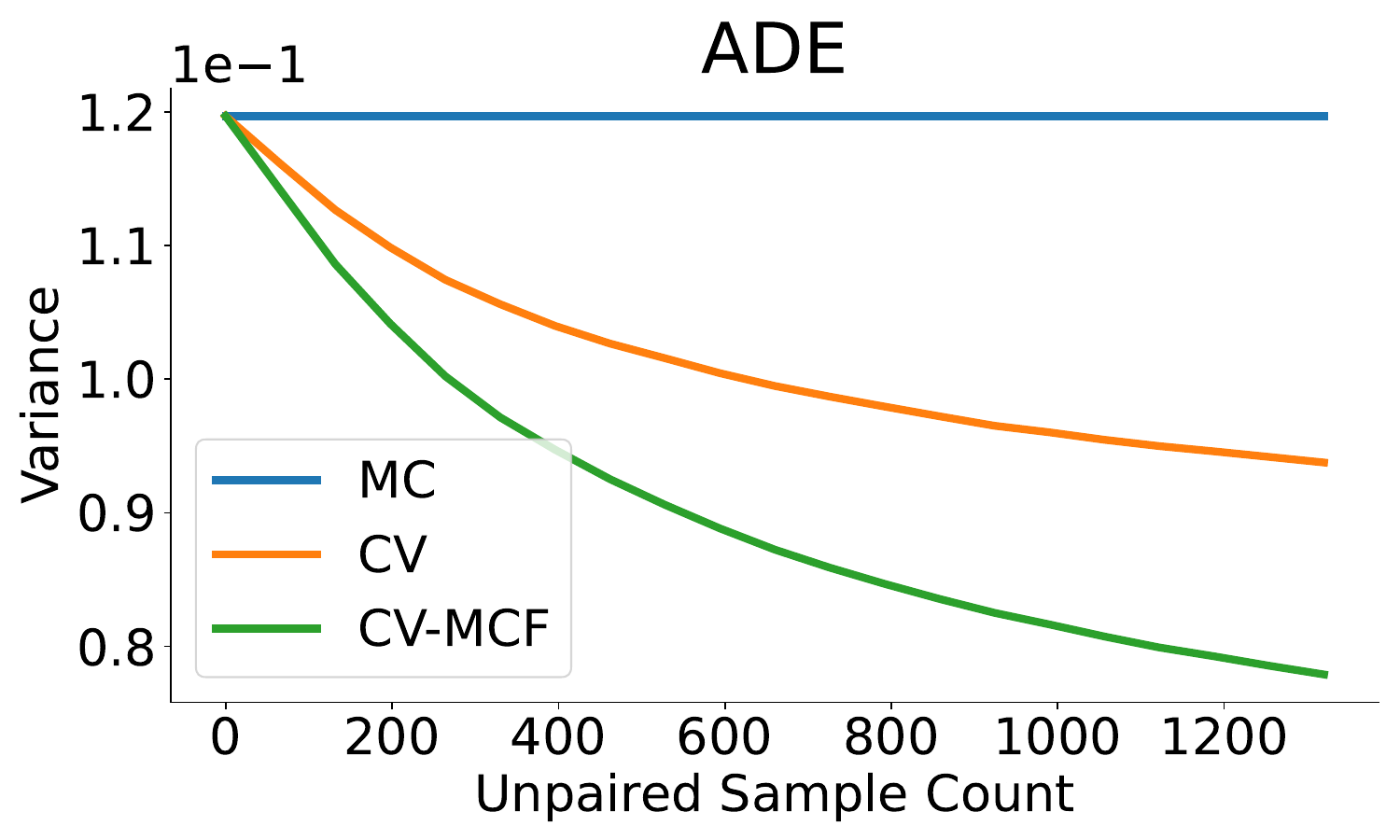}
    \end{subfigure}
    \begin{subfigure}{0.44\textwidth}
    \includegraphics[width=\textwidth]{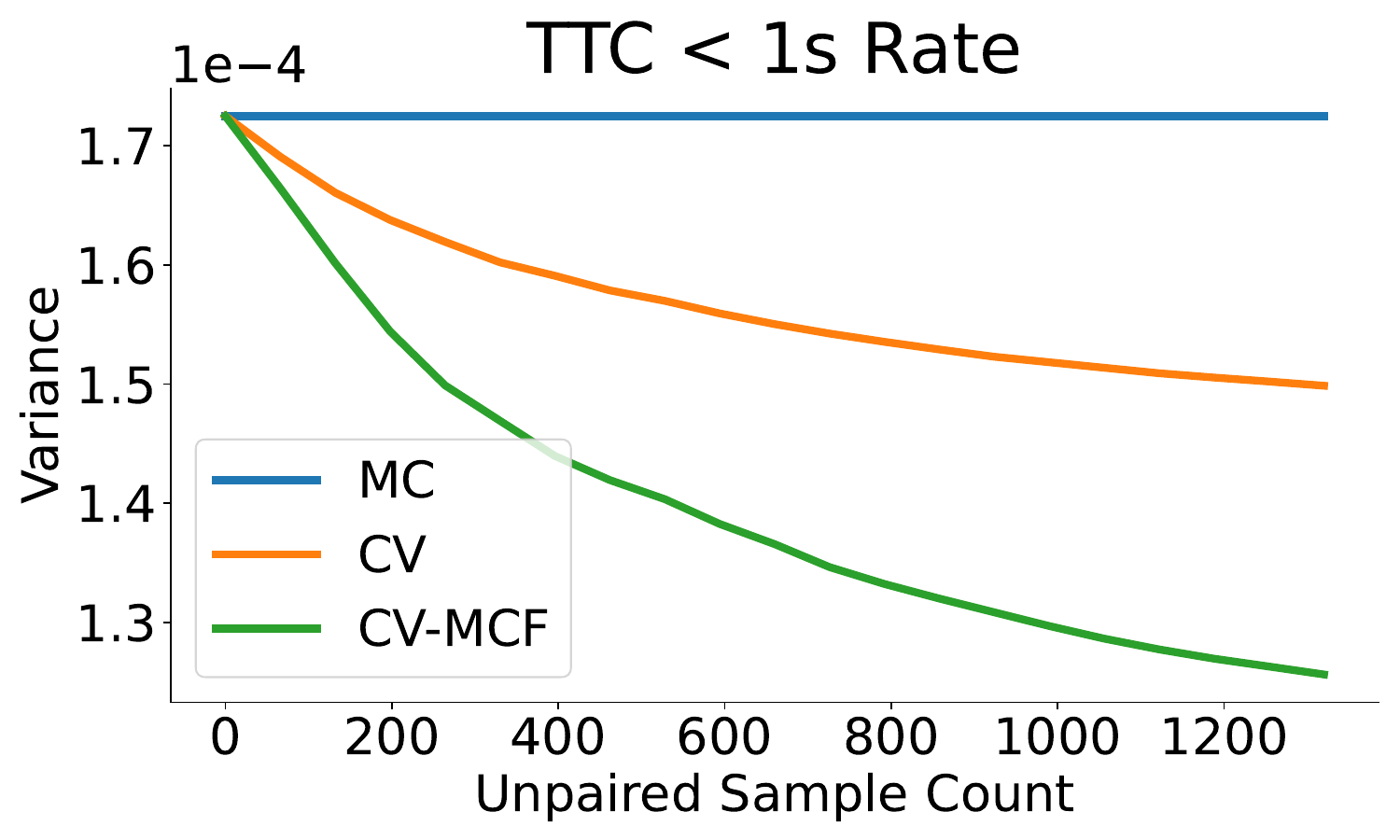}
    \end{subfigure}
    \caption{Variance estimates for two representative metrics as a function of the number of unpaired open-loop samples $\nsurrogate$ in the ``on\_intersection'' scenario class. Plots averaged over 10 random trials. }
    \label{fig:nuplan_k_variance}
    \vspace{-2mm}

\end{figure}

\vspace{-2mm}
\paragraph{Variance vs. number of unpaired samples.}
We first fix the number of paired samples $\npaired$ and sweep the number of open-loop only samples $\nsurrogate$. 
Figure~\ref{fig:nuplan_k_variance} plots the sample variance of each estimator for two representative metrics (ADE and TTC $<$ 1s rate) 
in the ``on\_intersection'' class, with $\npaired=565$ and $\nsurrogate$ from 0 to 1320. As $\nsurrogate$ increases, the control variate estimators rapidly reduce the variance relative to the Monte Carlo baseline -- achieving up to 22\% reduction for the direct approach and 35\% reduction for the learned approach when $\nsurrogate=1320$. Similar results for additional metrics and scenario classes are shown in the Appendix. 

\vspace{-2mm}
\paragraph{Impact of the learned metric correlator function.}
In settings where $\rho(\surrogate,\metric)$ is relatively low (e.g. $\rho(\surrogate,\metric) \approx 0.43$ for the time-to-collision $<$ 1s metric in the ``on\_intersection'' scenario class), directly using  $\surrogate$ as the control variate yields limited gains. We therefore train a lightweight feed-forward network $\mcf$ on a subset of the $\npaired$ paired samples, predicting $\metric$ from the set of all available open-loop metrics $\surrogate$ and additional scene features $\feat(\scenario)$; we augment the target domain samples with data from non-target domains (i.e. other scenario classes). While training the MCF may ``consume'' paired samples that could otherwise be used directly in the control variate computation, it can be worthwhile if the resulting correlation improvement compensates for the reduced sample size. See Appendix for additional training details.

Figure~\ref{fig:nuplan_variance_vs_n_train} plots the variance as a function of the fraction of in-domain paired samples used for training the MCF (with fixed $\nsurrogate$) for the ``on\_intersection'' scenario class, averaged over 10 trials.  Results for additional metrics and scenario classes are shown in the Appendix. 
We find that the improvement from the metric correlator function is more pronounced in domains with weaker baseline correlation.
Notably, even when the model is trained exclusively on out-of-domain data (i.e. zero in-domain samples are used), it still improves the in-domain correlation, indicating strong generalization from related scenarios. Crucially, this comes \textit{without} sacrificing any of the $\npaired$ samples used for MC or CV estimation, yielding consistent variance improvements across subdomains. 

\begin{figure} 
    \centering
    \begin{subfigure}[b]{0.44\textwidth}
        \centering
        \includegraphics[width=\textwidth]{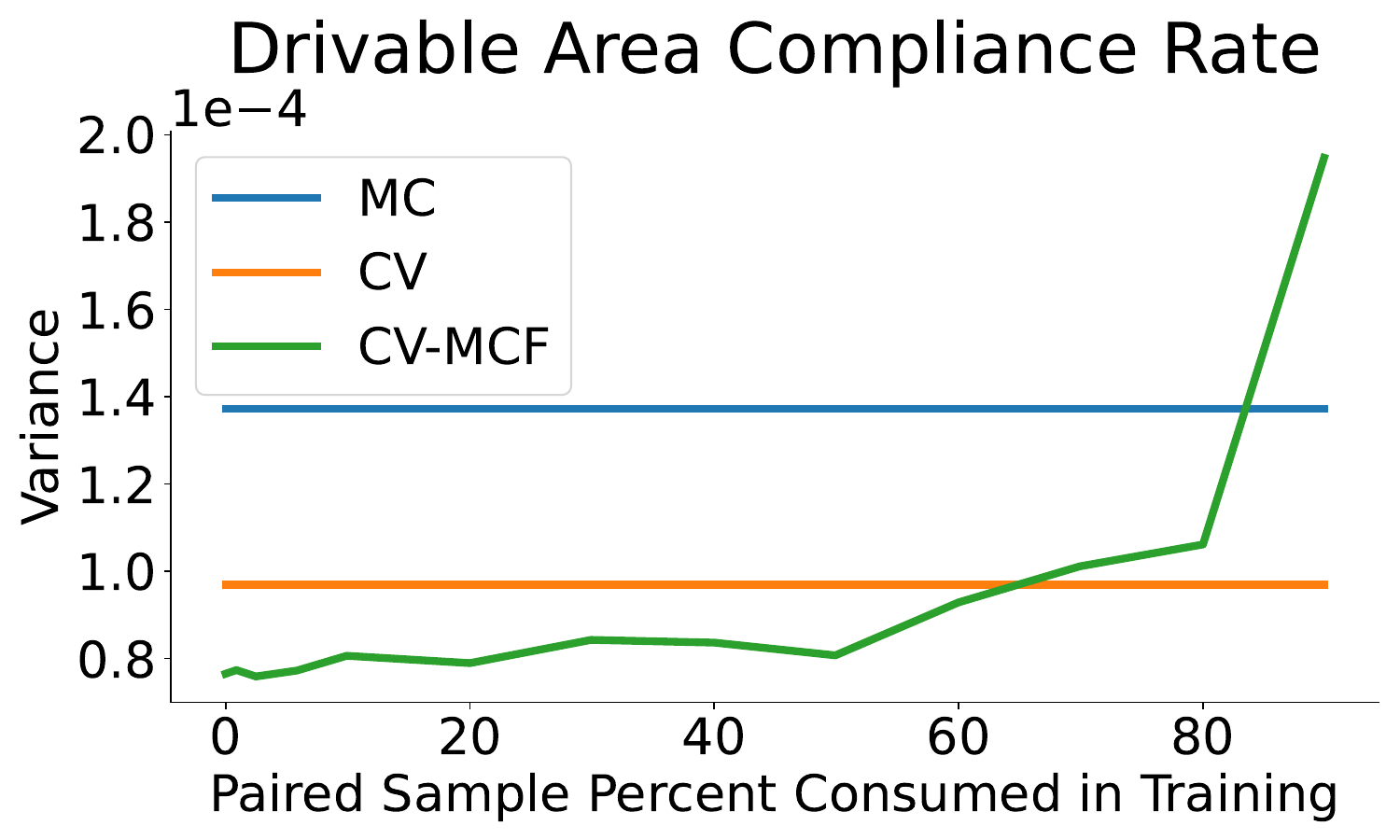}
    \end{subfigure}
    \begin{subfigure}[b]{0.44\textwidth}
        \centering
        \includegraphics[width=\textwidth]{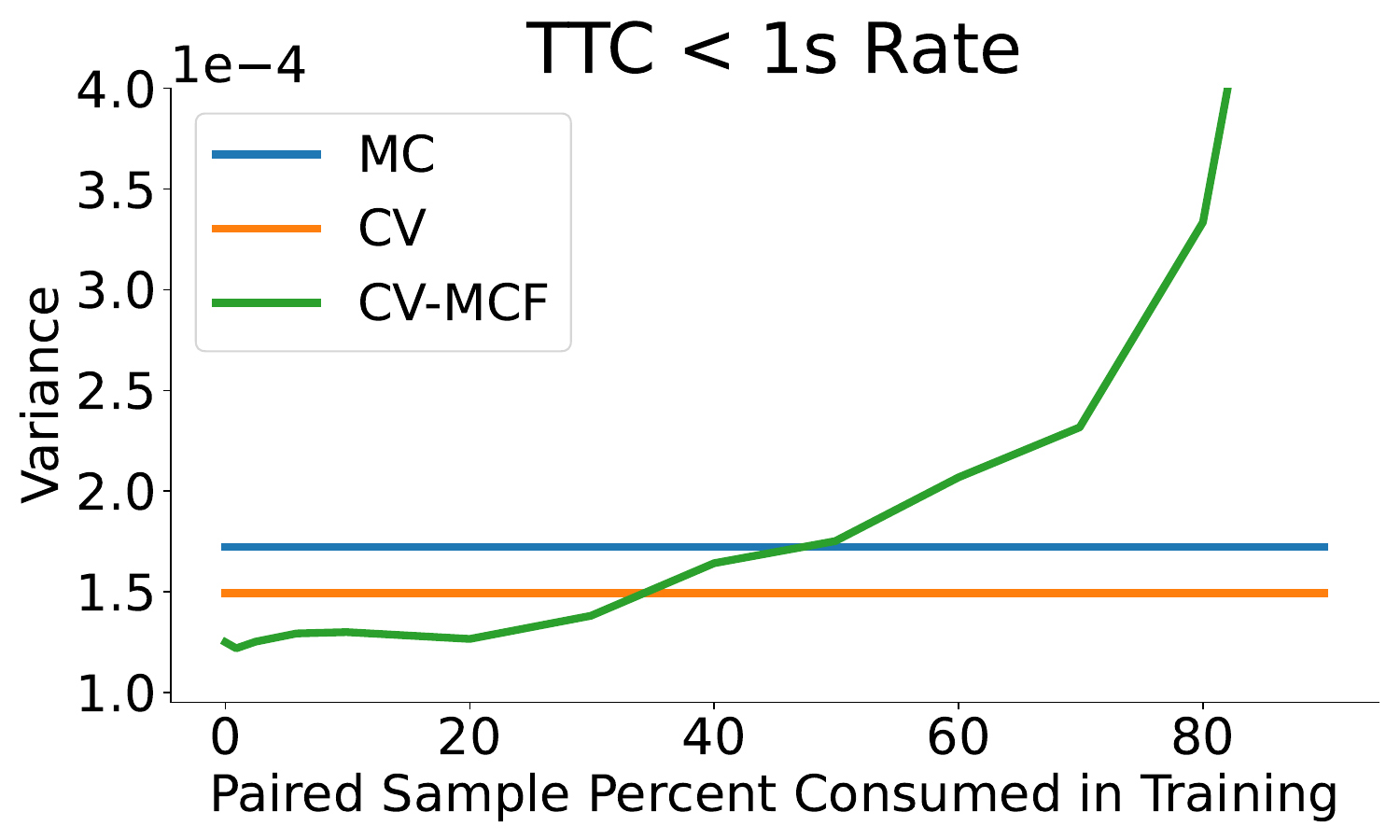}
    \end{subfigure}
    \caption{Variance estimates for two representative metrics as a function of the fraction of samples used to train the MCF in the ``on\_intersection'' scenario class. Using data to train an MCF to improve correlation can reduce variance, provided enough data remains for the control variates estimator.
    Plots averaged over 10 random trials.}
    \vspace{-4mm}
    \label{fig:nuplan_variance_vs_n_train}
\end{figure}   

Our experiments demonstrate that using open-loop metrics as control variates significantly reduces the variance in estimating closed-loop performance across various metrics and scenario classes. This variance reduction enables stronger statistical guarantees using fewer expensive closed-loop evaluations. As a representative example, by applying  Theorem~\ref{thm:sample-complexity} in the ``near\_long\_vehicle'' domain for the ADE metric with $n_r=715$, $\nsurrogate=1669$, $\rho_{CV}=0.79$, and $\rho_{CV-MCF}=0.83$, we find that the same confidence interval can be achieved with just $n_{p,\text{CV}} = 350$ (a 51\% reduction) or $n_{p,\text{CV-MCF}} = 303$ (a 58\% reduction) paired samples.

\subsection{Real-World Autonomous Driving Performance}

We next evaluate the effectiveness of our control variate estimator in assessing real-world autonomous driving performance in different types of scenarios (e.g. at road curves or at intersections). To this end, we use paired data comprising outputs from an internal, state-of-the-art neural reconstruction simulator and corresponding real-world driving logs, all collected under the same driving policy. The simulator is capable of reconstructing logged driving scenes (and can render novel camera views when the ego‐vehicle deviates from its original trajectory) \cite{wu2025alpamayo}, enabling the construction of paired (simulated, real-world) scenes. For each paired episode ($\npaired=138$ for intersection events and $\npaired=75$ for curvy road events), we extract two real‐world metrics $\metric$ --- nearest distance to another vehicle and average centerline deviation. We also obtain their simulated counterparts \(\surrogate\), and generate additional simulated samples ($\nsurrogate=781$ for intersection events and $\nsurrogate=423$ for curvy road events). Our objective is to leverage the relatively abundant and inexpensive simulator data to improve inference of real-world driving performance. Since the raw correlation \(\rho(\surrogate,\metric)\) exceeds 0.9 for both metrics, we apply the control variate estimator directly, without learning a metric correlator function.  
Table~\ref{tab:alpasim_vars} reports the sample variance of each estimator for the two metrics.  Relative to the Monte Carlo baseline, the control variate estimator achieves variance reductions of up to 82.9\%.

\begin{table}[h!]
    \centering
    \begin{tabular}{l|ccccc}     
        \toprule
         Class/Metric & \muhatmc & \muhatbetahat & 
         $\widehat{\Var}$( \muhatmc ) &  $\widehat{\Var}$( \muhatbetahat ) & Var Reduction\\ 
         \midrule
         intersections/closest dist & 10.5636 & 9.5850 & 1.0374 & 0.1776 & 82.9 \% \\
         curvy roads/closest dist & 10.7699 & 11.9377 & 1.4205 & 0.4335 & 69.5 \% \\
         intersections/lane centering dist & 0.7065 & 0.6917 & 0.0016 & 0.0003 & 81.7 \% \\
         curvy roads/lane centering dist & 0.6376 & 0.6647 & 0.0038 & 0.0010 & 74.3 \% \\
        \bottomrule
    \end{tabular}
    \vspace{1mm}
    \caption{Estimates of the mean and the variance of the mean for two different metrics, where ($n=138$, $k=781$) for intersection events and ($n=75$, $k=423$) for curvature events.}
    \label{tab:alpasim_vars}
    \vspace{-2mm}
\end{table}

These results confirm that our Sim2Val framework substantially tightens confidence intervals on real‐world performance estimates. For example, for intersection events with the closest distance metric, the Monte Carlo estimator requires $n_r=807$ real-world samples to match the confidence interval of the control variate estimator with $n_p=138$ real-world samples (where $k=781$ and $\rho_{cv}=0.995$), a nearly 6x increase in real-world data requirements.

\subsection{Quadruped Velocity Tracking}

We then consider the problem of evaluating the velocity tracking performance of a policy trained via reinforcement learning in simulation, and then deployed on a quadruped robot in the real world. Here, the scenario is parameterized by $X \in \mathbb{R}^3$, the three-dimensional command velocity vector, representing desired velocities in the forward, sideways, and yaw directions, respectively. The metric of interest $\metric$ here is the average velocity tracking error seen over a 2 second deployment of the reinforcement learning policy in the real world, while the performance of the policy in a rollout in the simulator serves as the surrogate metric $\surrogate$. We collect $\npaired=200$ paired samples from real-world tests and paired simulation results, and $\nsurrogate=400$ additional simulator-only samples.   

\begin{table}[h]
    \centering
    \begin{tabular}{c|cccc}
        \toprule
        Method & MC & CV & CV-MCF ($\nfit=50$) & CV-MCF ($\nfit=40$) \\
        \midrule
        Mean & $4.815 \times 10^{-1}$ & $4.811 \times 10^{-1}$ & $4.748 \times 10^{-1} $  & $4.749 \times 10^{-1} $ \\
        Variance & $2.048 \times 10^{-5}$ & $2.041 \times 10^{-5}$ & $1.931 \times 10^{-5}$ & $1.926 \times 10^{-5}$\\
        \bottomrule
    \end{tabular}
    \vspace{1mm}
    \caption{Estimated means and variances using different methods.}
    \label{tab:quadruped_summary}
\end{table}
\vspace{-2mm}

\textbf{Direct control variates.} 
We first apply the direct control variates approach (CV). As shown in Table~\ref{tab:quadruped_summary}, the direct CV method reduces the variance from $2.048 \times 10^{-5}$ (using only the $n$ real-world evaluations) to $2.041 \times 10^{-5}$. The Pearson correlation coefficient $\rho$ between the $n$ paired simulation and real-world metrics is $0.0728$, indicating a weak correlation between simulated and real-world performance; consequently, the reduction in variance achieved is limited. By applying Eq.~\ref{eq:cv-data-reduction} from Theorem \ref{thm:sample-complexity}, we compute $n_{\min}=200$, indicating that when $\rho$ is close to zero, the reduction in sample complexity is negligible.

\textbf{Control variates with learned metric correlator function.} 
Given the weak correlation between the simulated and real-world metrics, we learn a mapping, conditioned on the command velocity $X$, that predicts the real-world metric from the simulation metric using $\nfit < \npaired$ paired evaluations. We then apply the control variates method with this learned mapping (CV-MCF).  
As shown in Table~\ref{tab:quadruped_summary}, when using $\nfit=50$ and $\nfit=40$ paired samples, respectively, the CV-MCF method further reduces the variance to $1.931 \times 10^{-5}$ and $1.926 \times 10^{-5}$. When $\nfit=50$, the Pearson correlation coefficient after learning the metric correlation function increases to $0.6158$, in which case $n_{\min} = 145 < \npaired$ from \eqref{eq:cv-data-reduction}, showing improved sample efficiency. Note that even this relatively modest reduction in variance from $2.048 \times 10^{-5}$ to $1.926 \times 10^{-5}$, would have required 38\% more real-world tests to achieve without leveraging simulation.


\section{Conclusion}
\label{sec:conclusion}

In this work, we present Sim2Val, a general, theoretically-grounded framework for improving metric estimation in robotics systems by leveraging \textit{paired} observations across heterogeneous test platforms. By treating auxiliary measurements -- such as simulator outputs or offline policy logs -- as control variates in a Monte Carlo estimator, our method provably decreases estimator variance whenever the auxiliary and target domains exhibit nontrivial correlation. We provide theoretical analysis on variance reduction and sample efficiency improvements, and empirically validate our approach in autonomous driving and quadruped robotics settings. This framework suggests several interesting avenues for future work, including developing policy-conditional extensions that adapt to evolving control policies, advancing distributional validation to encompass quantile estimation beyond the mean, and optimally allocating repeated simulator calls under a fixed compute budget to further reduce estimator variance.


\clearpage

\section{Limitations}
\label{sec:limitations}
Our theoretical analysis relies on the assumption that all test samples (real and surrogate) are drawn i.i.d. from the target distribution. In practice, data collection is often biased towards ``interesting'' or safety-critical scenarios. Addressing this mismatch may require integrating selection biases into our framework, for example by viewing targeted sampling as a form of importance sampling to further reduce variance. Additionally, by focusing exclusively on mean performance, our approach may not fully capture safety-critical concerns, where uniform reliability across diverse operating conditions is important. Although conditioning on specific operational design domains can reveal deficiencies masked by the population mean, extending our methodology to tail-risk measures (such as conditional value at risk or other quantile-based metrics) would be a useful direction for future work. Finally, we have assumed that only a single auxiliary sample $\surrogate \sim \surrogatedist$ is available per scenario $\scenario$, despite the fact that simulators can generate multiple roll-outs of $\surrogate$ from the same $\scenario$. Optimally allocating repeated simulator calls across scenarios could yield additional variance reduction in the control variate estimator for $\mu$, and would be an interesting avenue for future work. 

\section*{Acknowledgment}
\label{sec:acknowledgment}
We thank Rajiv Swamy and Alex Tong for their efforts in collecting the quadruped dataset in the Computational Robotics Lab at Harvard University.


\bibliography{reference}  

\newpage
\input{appendix}

\end{document}

%% file: appendix.tex
\appendix
\section{Proofs}
\label{app:proofs}

\begin{proof}[\textbf{Proof of Theorem~\ref{thm:variance-reduction}}]
We start by deriving the variance of the control variate estimator for fixed $\coeff$. 
\begin{align}
    \Var(\hat\mu_\mathrm{CV}) &= \frac{1}{\npaired} \Var( \metric - \coeff^T \surrogate) + \frac{1}{\nsurrogate} \Var( \coeff^T \surrogate) \nonumber \\
    &= \frac{1}{\npaired} \left( \Var( \metric )  - 2 \coeff^T \Cov(\surrogate, \metric) + \coeff^T \Var (\surrogate) \coeff) \right) + \frac{1}{\nsurrogate} \coeff^T \Var( \surrogate) \coeff \nonumber \\
    &= \frac{1}{\npaired} \Var(\metric) - \frac{2}{\npaired} \coeff^T \Cov(\surrogate,\metric) + \frac{\npaired + \nsurrogate}{\npaired \nsurrogate} \coeff^T \Var(\surrogate) \coeff \label{eq:var-cov-1}
\end{align} 

Noting that $\Var(\hat\mu_\mathrm{CV})$ is a convex quadratic in $\coeff$, we determine the optimal $\coeff$ that minimizes this variance by differentiating \eqref{eq:var-cov-1} with respect to $\coeff$ and setting the result equal to zero, which gives:
\begin{align*}
    \bm{0} &= - \frac{2}{\npaired} \Cov(\surrogate,\metric) - 2 \frac{\npaired + \nsurrogate}{\npaired \nsurrogate} \Var(\surrogate) \coeff_\mathrm{opt}  \\
    \coeff_\mathrm{opt} &= \frac{\nsurrogate}{\npaired +\nsurrogate} \Var( \surrogate )^{-1} \Cov(\surrogate,\metric).
\end{align*}

To obtain the optimal variance, we plug this result back into \eqref{eq:var-cov-1}:

\begin{align*}
    \Var(\hat\mu_\mathrm{CV}) &= \frac{1}{\npaired} \Var(\metric) \\
    &\hspace{1cm} - \frac{2}{\npaired} \frac{\nsurrogate}{\npaired +\nsurrogate} \Cov(\metric,\surrogate) \Var( \surrogate )^{-1}  \Cov(\surrogate,\metric) \\
    &\hspace{1cm} + \frac{\npaired + \nsurrogate}{\npaired \nsurrogate} \left( \frac{\nsurrogate}{\npaired +\nsurrogate} \right)^2  \Cov(\metric,\surrogate) \Var( \surrogate )^{-1} \Var(\surrogate)  \Var( \surrogate )^{-1}  \Cov(\surrogate,\metric) \\
    &= \frac{1}{\npaired} \left( \Var(\metric) - \frac{\nsurrogate}{\npaired +\nsurrogate} \Cov(\metric,\surrogate) \Var( \surrogate )^{-1}  \Cov(\surrogate,\metric) \right) \\
    &= \frac{1}{\npaired} \bigg( 1 -  \frac{\nsurrogate}{\npaired+\nsurrogate} \underbrace{ \Cov(\metric,\surrogate) \Var( \surrogate )^{-1}  \Cov(\surrogate,\metric) \Var(\metric)^{-1} }_{:=\rho^2(\surrogate,\metric)} \bigg) \Var(\metric).
\end{align*}
We can simplify $\rho^2(\surrogate,\metric)$ as
\begin{align*}
    \rho^2(\surrogate,\metric) &= \Cov(\metric,\surrogate) \Var( \surrogate )^{-1}  \Cov(\surrogate,\metric) \Var(\metric)^{-1} \\
    &= \mathrm{Tr}\left(\Cov(\metric,\surrogate) \Var( \surrogate )^{-1}  \Cov(\surrogate,\metric) \Var(\metric)^{-1} \right) \\
    &= \mathrm{Tr}\left( \Var(\metric)^{-\frac12} \Cov(\metric,\surrogate) \Var( \surrogate )^{-\frac12} \Var( \surrogate )^{-\frac12}\Cov(\surrogate,\metric) \Var(\metric)^{-\frac12} \right) \\
    &= \left\| \Var(\surrogate)^{-\frac12} \Cov(\surrogate,\metric) \Var( \metric )^{-\frac12} \right\|^2_\mathrm{Fr},
\end{align*}
where the first step is possible since $\rho^2(\surrogate,\metric)$ is a scalar (since $\metric$ is a scalar).
\end{proof}

\begin{proof}[\textbf{Proof of Theorem~\ref{thm:sample-complexity}}]
For any unbiased estimator $\hat\mu(\mathcal{D})$ of $\mu := \mathbb{E}[\metric]$ with variance $\Var(\hat\mu)$, Chebyschev's inequality states that
\begin{align}
    \mathbb{P}\bigg( | \hat\mu - \mu | \geq \alpha \bigg) \leq \frac{\Var(\hat\mu)}{\alpha^2}. \label{eq:chebyschev-gen}
\end{align}

Fixing a desired confidence level $1-\delta$, the confidence interval over $\mu$ is,
\begin{align}
    \hat\mu \pm \sqrt{\frac{\Var(\hat\mu)}{\delta}}, \label{eq:chebyschev-conf-gen}
\end{align}
i.e., the confidence interval has a radius of $\sqrt{\frac{\Var(\hat\mu)}{\delta}}$.

Let $n_r\in\mathbb{Z}_{>0}$ be the number of real-world samples of $\metric$ for computing $\hat{\mu}_\mathrm{MC}$ (as in Eq.~\ref{eq:mc_estimator}). Let $n_p\in\mathbb{Z}_{>0}$ be the number of real-world and surrogate data pairs $(\metric,\surrogate)$ drawn for computing $\hat\mu_\mathrm{CV}$ using control variates (as in Eq.~\ref{eq:cv-estimator} and~\ref{eq:cv_beta_opt}). 

First, we consider the case where $n_r = n_p = n$. Here, we have
\begin{align*}
    \Var(\hat\mu_\mathrm{CV}) &= \frac{\Var(\metric)}{n}\left( 1 - \frac{k}{k+n}\rho^2(\surrogate, \metric) \right) \\
    &\le \frac{\Var(\metric)}{n} = \Var(\hat\mu_\mathrm{MC}) 
\end{align*}
since $\rho^2(\surrogate,\metric) \in [0,1]$. 
Plugging these into Eq. \ref{eq:chebyschev-conf-gen}, we see that the consequence of this is that the confidence interval of the control variate estimator will be tighter than that of the simple Monte Carlo estimator.

Next, we fix a desired interval size $\alpha$. To achieve a desired confidence in this interval of $1-\delta$, we need
\begin{align*}
    \delta = \frac{\Var(\hat\mu)}{\alpha^2}.
\end{align*}
For the MC estimator, we have
\begin{align}
    \delta &= \frac{1}{\alpha^2} \frac{\Var(\metric)}{n_r} \\
    n_r &= \frac{\Var(\metric)}{\alpha^2 \delta}. \label{eq:num-real-only}
\end{align}
For the control variate estimator, we have
\begin{align}
    \delta &= \frac{1}{\alpha^2} \frac{\Var(\metric)}{n_p}\left( 1 - \frac{k}{k+n_p}\rho^2(\surrogate, \metric) \right)  \\
    n_p \delta &= \frac{\Var(\metric)}{\alpha^2} - \frac{k \Var(\metric)}{(k + n_p) \alpha^2}\rho^2(\surrogate,\metric) \\
    n_p (k + n_p)&= (k + n_p) \frac{\Var(\metric)}{\alpha^2 \delta} + \frac{k \Var(\metric)}{\alpha^2 \delta}  \rho^2(\surrogate,\metric). \label{eq:num-cv}
\end{align}
For a given confidence interval specified by $\alpha$ and $\delta$, Eq. \ref{eq:num-real-only} relates $\Var(\metric)$, $\alpha$, and $\delta$ to the number of samples required using simple Monte Carlo. To compute the minimum number of samples required to achieve the same interval with the control variate estimator, we substitute Eq.~\ref{eq:num-real-only} into Eq.~\ref{eq:num-cv} above, and replace $n_p$ with $n_\mathrm{min}$
\begin{align}
     n_\mathrm{min} (k + n_\mathrm{min}) &= (k + n_\mathrm{min}) n_r + k n_r \rho^2(\surrogate,\metric) \\
     n_\mathrm{min} k + n_\mathrm{min}^2 &= n_r k + n_\mathrm{min} n_r + k n_r \rho^2(\surrogate,\metric). 
\end{align}
Solving this quadratic for $n_\mathrm{min}$, we have
\begin{align}
    n_\mathrm{min} &= \frac12 \left( - (k-n_r) + \sqrt{(k-n_r)^2 + 4 n_r k ( 1 - \rho^2(\surrogate,\metric) ) } \right), 
\end{align}
which gives us the required number of paired samples such that $\hat\mu_\mathrm{CV}$ achieves the same confidence interval as $\hat\mu_\mathrm{MC}$. 

Finally, we show that $n_p \in [0, n_r]$. We begin by noting that $n_\mathrm{min}$ is monotonically decreasing in $\rho^2$. Hence, setting $\rho^2=0$, we attain the maximum value of $n_\mathrm{min} = n_r$, while setting $\rho^2 = 1$, we attain the minimum value $n_\mathrm{min} = 0$. As $n_p$ is a continuous monotonic function in $\rho^2$, it will attain all values between its extrema of $0$ and $n_r$.
\end{proof}

\section{Computing Sample Means and Variances}

We computed $\hat\mu$ and $\widehat{\mathrm{Var}}$ in our experiments as follows.

The plug-in estimator for $\beta_{opt}$ (a scalar in our experiments) is
\[
\hat\beta_{opt}
= \Bigl(\frac{k}{k+n}\Bigr) \frac{\displaystyle\sum_{i=1}^n \bigl(\metric_i-\bar \metric\bigr)\,\bigl(G_i-\bar G\bigr)}
       {\displaystyle\sum_{i=1}^n \bigl(G_i-\bar G\bigr)^2},
\]
where
\[
\bar \metric = \frac{1}{n}\sum_{i=1}^n \metric_i,
\quad
\bar G = \frac{1}{n}\sum_{i=1}^n G_i.
\]

The plug-in estimator for $\hat\mu_\beta$ is
\[
\hat\mu_{\hat\beta} = \frac{1}{n}\sum_{i=1}^n\bigl(\metric_i - \hat\beta\,G_i\bigr) + \hat\beta\,\hat\theta,
\]
where
\[
\hat \theta = \frac{1}{k}\sum_{j=1}^k G'_j.
\]

The sample variance of $\hat\mu_{\hat\beta}$ is then
\[
\widehat{\mathrm{Var}}(\hat\mu_{\hat\beta}) = \frac{1}{n(n-1)}\sum_{i=1}^n\bigl(\metric_i - \hat\beta\,G_i - \bar \metric + \hat\beta\,\bar G \bigr)^2 + \frac{\hat\beta^2}{k(k-1)}\sum_{j=1}^k\bigl(G'_j - \hat\theta)^2.
\]

Note that using the same data to estimate $\beta_{opt}$ and to compute the control variate estimator introduces some bias; this is standard with control variates when $\beta_{opt}$ is estimated from the same samples. However, this bias is typically quite small in practice -- it is $O(1/n)$ and dominated by the $O(1/\sqrt{n})$ error between the estimator and the true mean \citep{owen2013monte}. If an unbiased estimate is needed, a simple approach is to estimate $\beta_{opt}$ from separate data, such as held-out samples or data from prior policy evaluations (ensuring that none of the current policy's samples are used). This allows $\beta_{opt}$ to be selected independently of the control variate estimate when $n$ is very small and bias is a concern. 

\section{Additional Experimental Details and Results}
\label{app:addl_exp_results}

\subsection{nuPlan Simulation Experiments}
\label{app:nuplan_results}

We generated paired open‐loop and closed‐loop driving data using the nuPlan simulator, instantiated with the Intelligent Driver Model (IDM) planner \citep{Treiber2000CongestedTS}.  For each scene, an open‐loop rollout records surrogate metrics under non‐reactive agent behavior, while a corresponding closed‐loop rollout -- with the same planner and policy -- captures metrics under reactive agent behavior.

The metric correlator function (MCF) is implemented as a fully connected neural network with three hidden layers. Each hidden layer applies a ReLU activation. The final output layer uses a sigmoid activation to produce a prediction bounded in \([0,1]\) for binary metrics.  The network input comprises
\[
  \surrogate_{i, train} = (\mathrm{ADE},\ \mathrm{TTC}{<}1\mathrm{s},\ \mathrm{drivable\_area\_compliance},\ \mathrm{no\_ego\_at\_fault\_collisions}),
\]
augmented by a feature vector \(\feat(\cdot)\) extracted from a trained trajectory prediction model~\citep{ding2025surprise}. The MCF maps \((\surrogate_{i, train},\ \feat)\) to a scalar estimate of the target closed‐loop simulation metric for the corresponding sample.

The MCF was trained on \(\nfit\) paired samples from our target domain (for instance the ``on\_intersection,'' ``near\_multiple\_vehicles,'' or ``near\_high\_speed\_vehicle'' scenario class), with a mean squared error loss for continuous metrics, and a binary cross-entropy loss for binary metrics (with early stopping based on validation loss to prevent overfitting). In some experiments, the target domain data was supplemented with paired samples from auxiliary domains (data from non-target scenario classes), and in others, the MCF was trained exclusively on target domain pairs.

\begin{figure} 
    \centering
    \begin{subfigure}{0.49\textwidth}
    \includegraphics[width=\textwidth]{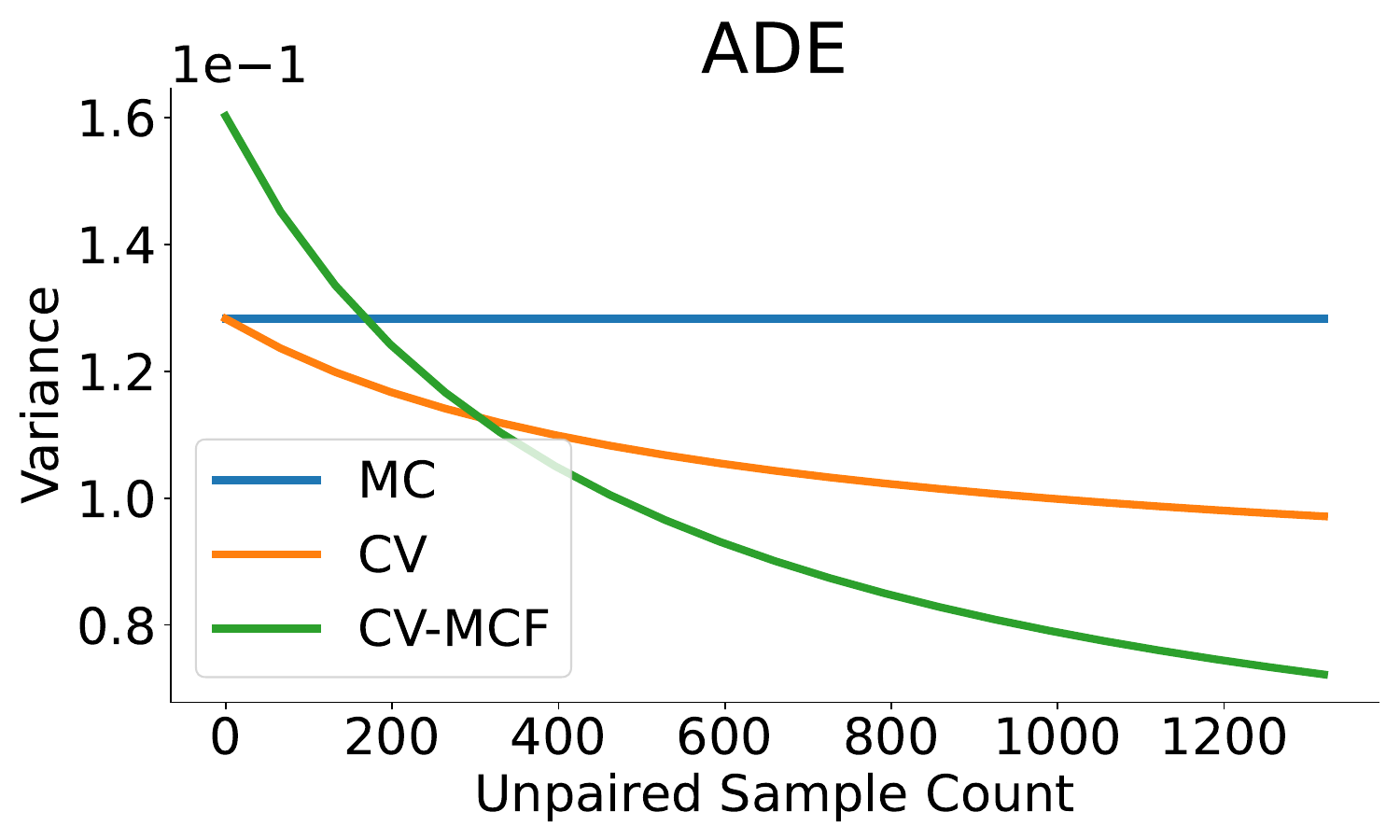}
    \caption{ADE metric, ``on\_intersection'' scenario class. MCF trained on in-domain pairs only. $\npaired=565$, $\nfit=113$.}
        \vspace{0.1in}
    \end{subfigure}
    \hfill
    \begin{subfigure}{0.49\textwidth}
    \includegraphics[width=\textwidth]{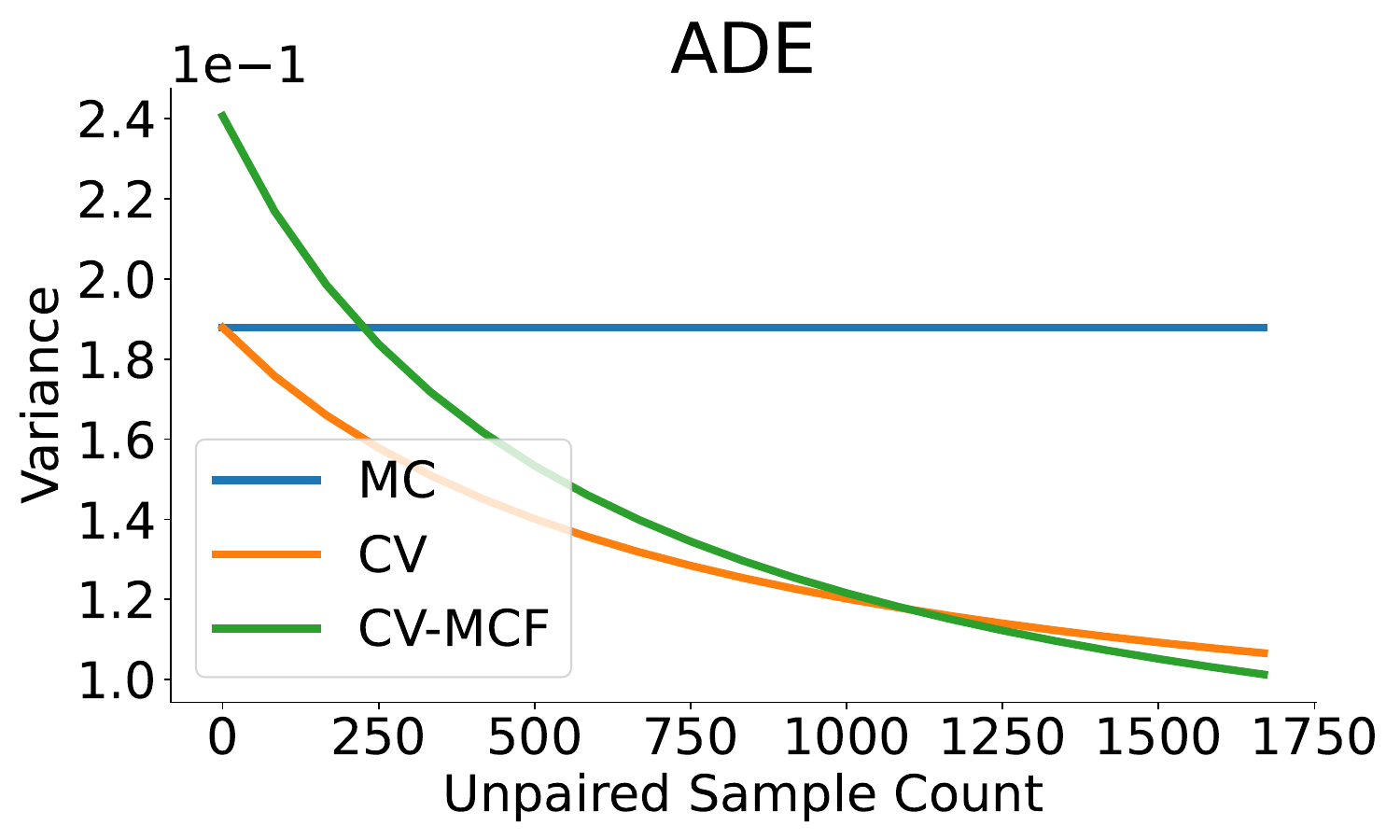}
    \caption{ADE metric, ``near\_long\_vehicle'' scenario class. MCF trained on in-domain pairs only. $\npaired=715$, $\nfit=143$.} 
    \vspace{0.1in}
    \end{subfigure}
    \begin{subfigure}{0.49\textwidth}
    \includegraphics[width=\textwidth]{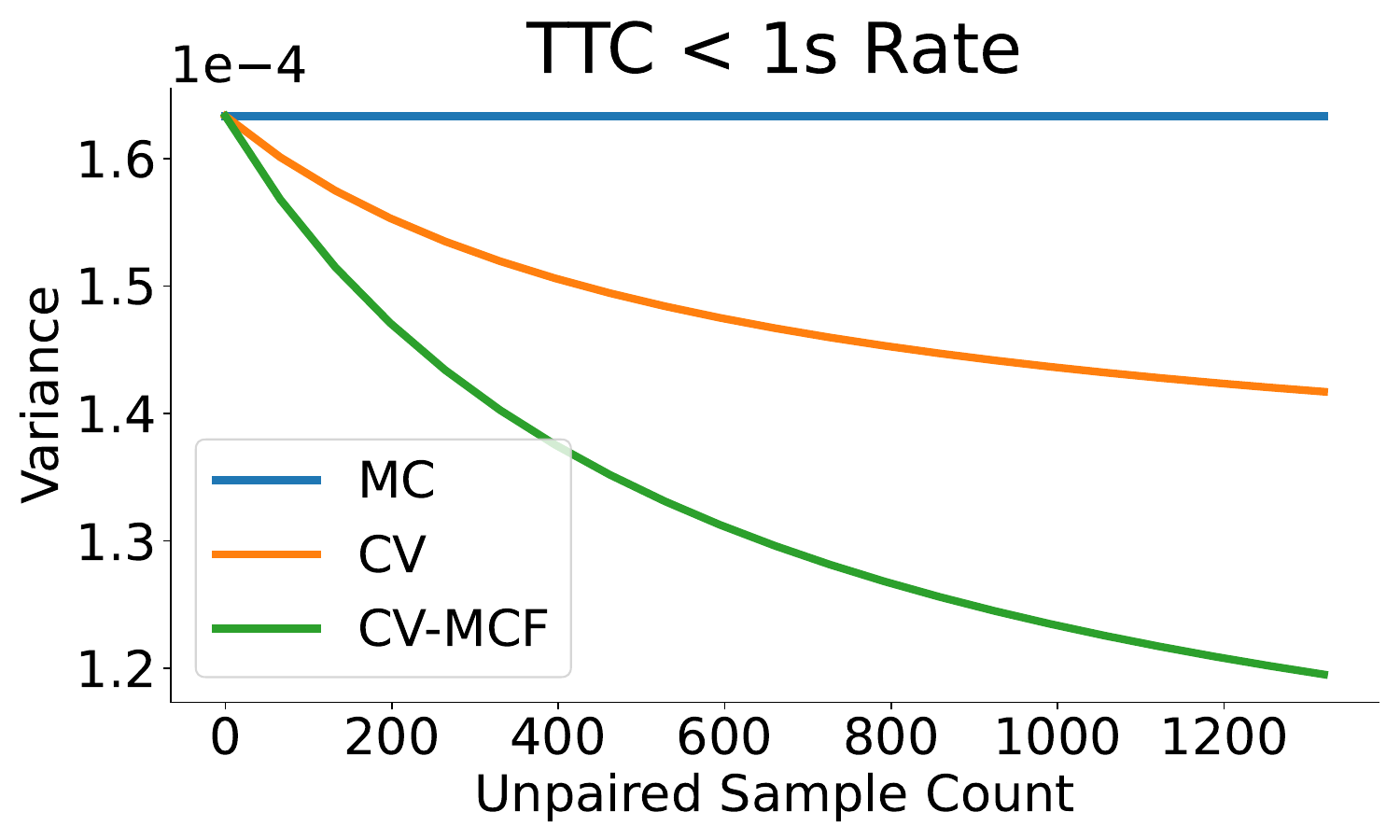}
    \caption{TTC $<$ 1s metric, ``near\_multiple\_vehicles'' scenario class. MCF trained on out-of-domain data only. $\npaired=432$, $\nfit=0$ (+ \text{out-of-domain data})}
    \end{subfigure}
    \hfill 
    \begin{subfigure}{0.49\textwidth}
    \includegraphics[width=\textwidth]{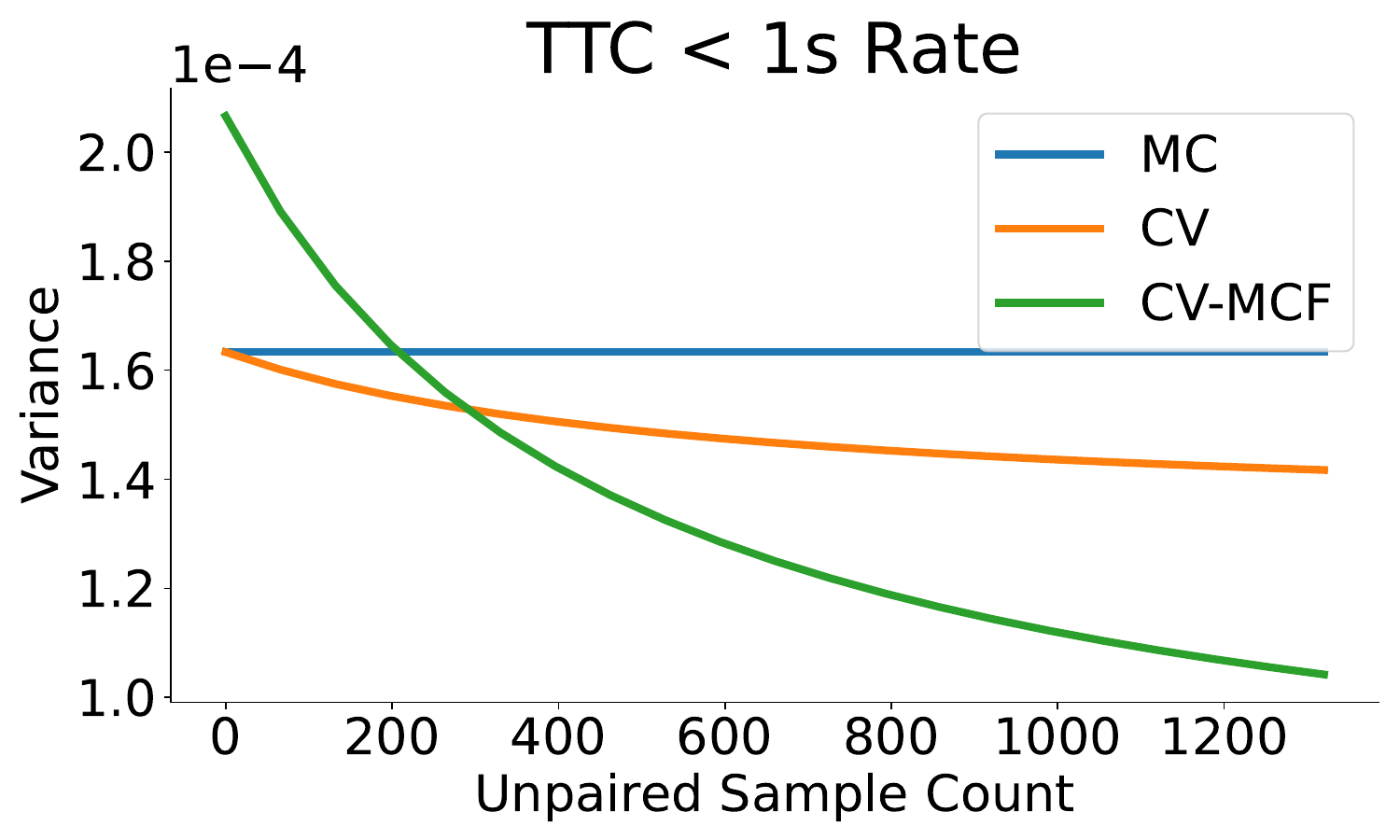}
    \caption{TTC $<$ 1s metric, ``on\_intersection'' scenario class. MCF trained on data from all domains. $\npaired=565$, $\nfit=113$ (+ \text{out-of-domain data})}
    \end{subfigure}
    \vspace{2mm}
    \caption{Variance vs. unpaired sample count for various metrics and domains. All plots averaged over 10 random trials. }
    \label{fig:nuplan_k_variance_appendix}
\end{figure}

\begin{figure} 
    \centering
    \begin{subfigure}{0.49\textwidth}
    \includegraphics[width=\textwidth]{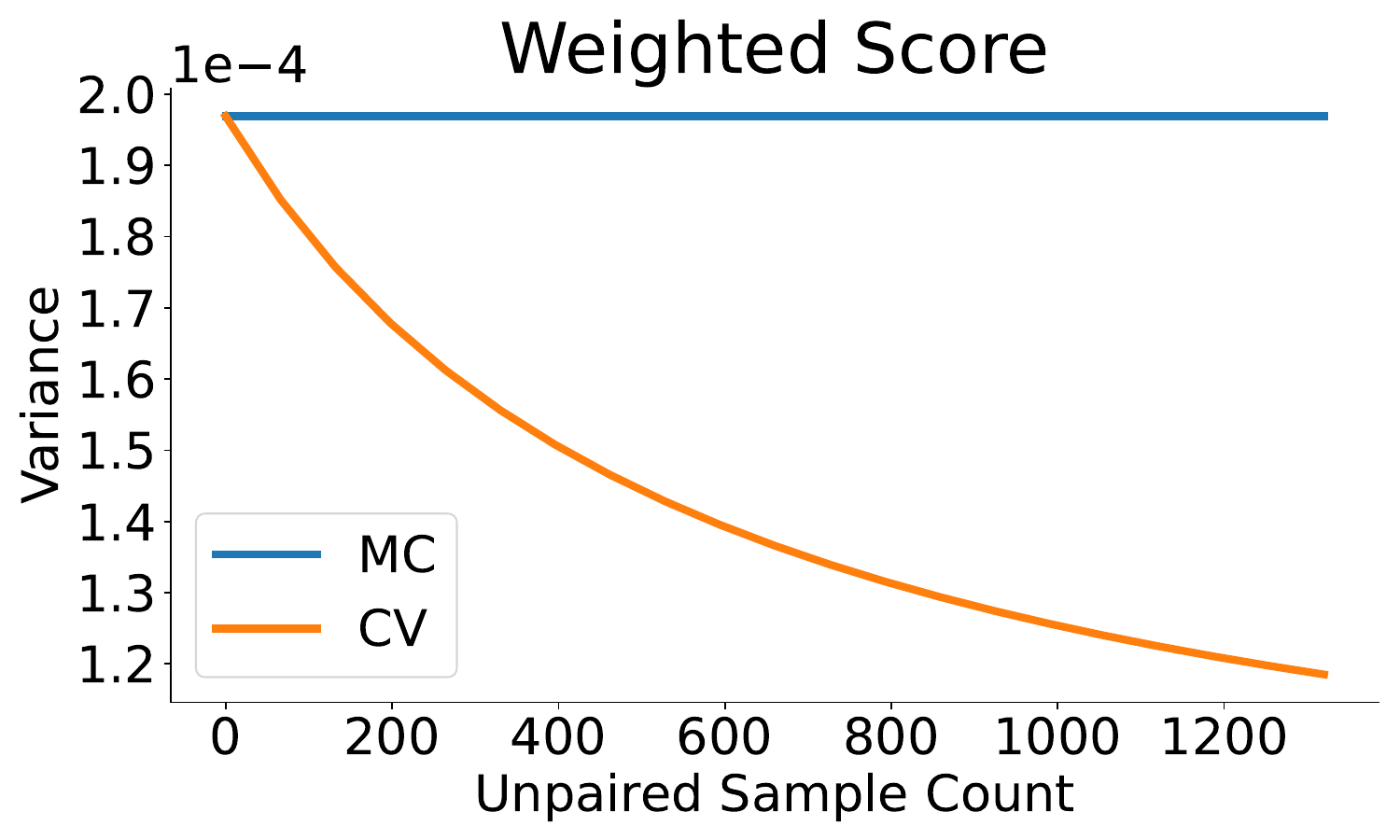}
    \caption{Aggregate score, ``on\_intersection'' class, $\npaired=565$, $\rho^2(\surrogate,\metric)=0.568$.}
        \vspace{0.1in}
    \end{subfigure}
    \hfill
    \begin{subfigure}{0.49\textwidth}
    \includegraphics[width=\textwidth]{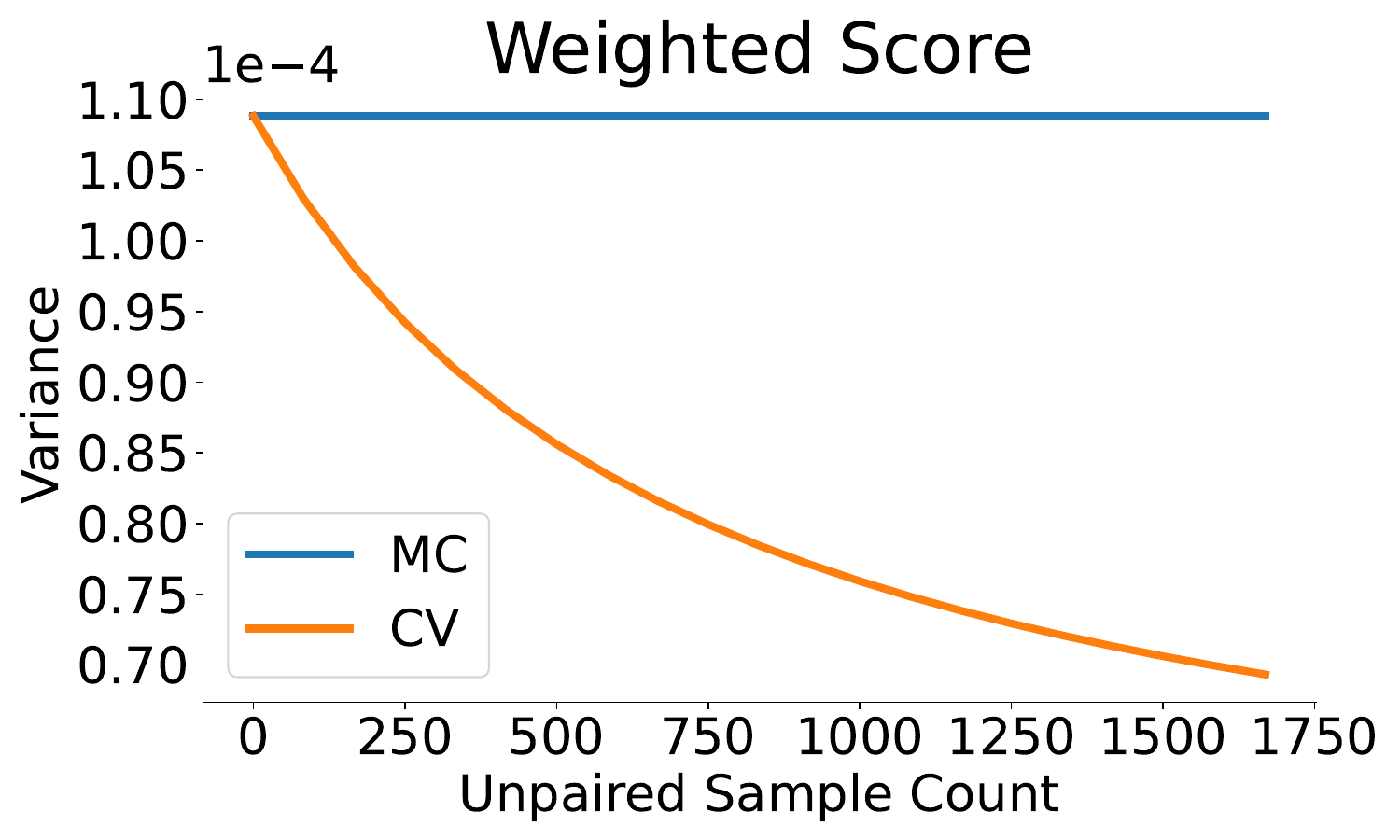}
    \caption{Aggregate score, ``near\_long\_vehicle'' class. $\npaired=715$, $\rho^2(\surrogate,\metric)=0.515$.} 
    \vspace{0.1in}
    \end{subfigure}
    \caption{Variance vs. unpaired sample count for NuPlan aggregate score, using ADE, TTC, drivable area compliance, and no ego at fault collisions as a control variate vector. Plots averaged over 10 random trials. }
   \label{fig:nuplan_k_variance_vector_appendix}
\end{figure}

Fig.~\ref{fig:nuplan_k_variance_appendix} compares the Monte Carlo (MC), control variates (CV), and control variates with MCF (CV–MCF) estimators across multiple metrics and domains as a function of the number of unpaired surrogate samples $k$. In all cases, estimator variance decreases as \(\nsurrogate\) increases. Panels (a) and (b) show results for models trained solely on target-domain pairs. In (a), the improved correlation from CV-MCF is high enough to outpace the CV approach, but in (b) it is not (until the unpaired sample count is much higher). Panels (c) and (d) correspond to models trained with auxiliary out-of-domain data pairs. Models trained on the out-of-domain data can generalize well to the in-domain data, resulting in lower variance estimates. Fig.~\ref{fig:nuplan_k_variance_vector_appendix} also plots the variance vs. the number of unpaired surrogate samples $k$, but uses the control variate vector (as in Eq.~\ref{eq:cv-variance}) rather than a scalar, with $\surrogate = (\text{ADE}, \text{TTC}, \text{drivable\_area\_compliance}, \text{no\_ego\_at\_fault\_collisions})$ and $\metric = \text{aggregate\_score}$. (The aggregate score is a weighted average of several individual metrics, and serves as an overall measure of driving performance.)

\begin{figure} 
    \centering
    \begin{subfigure}[t]{0.49\textwidth}
    \includegraphics[width=\textwidth]{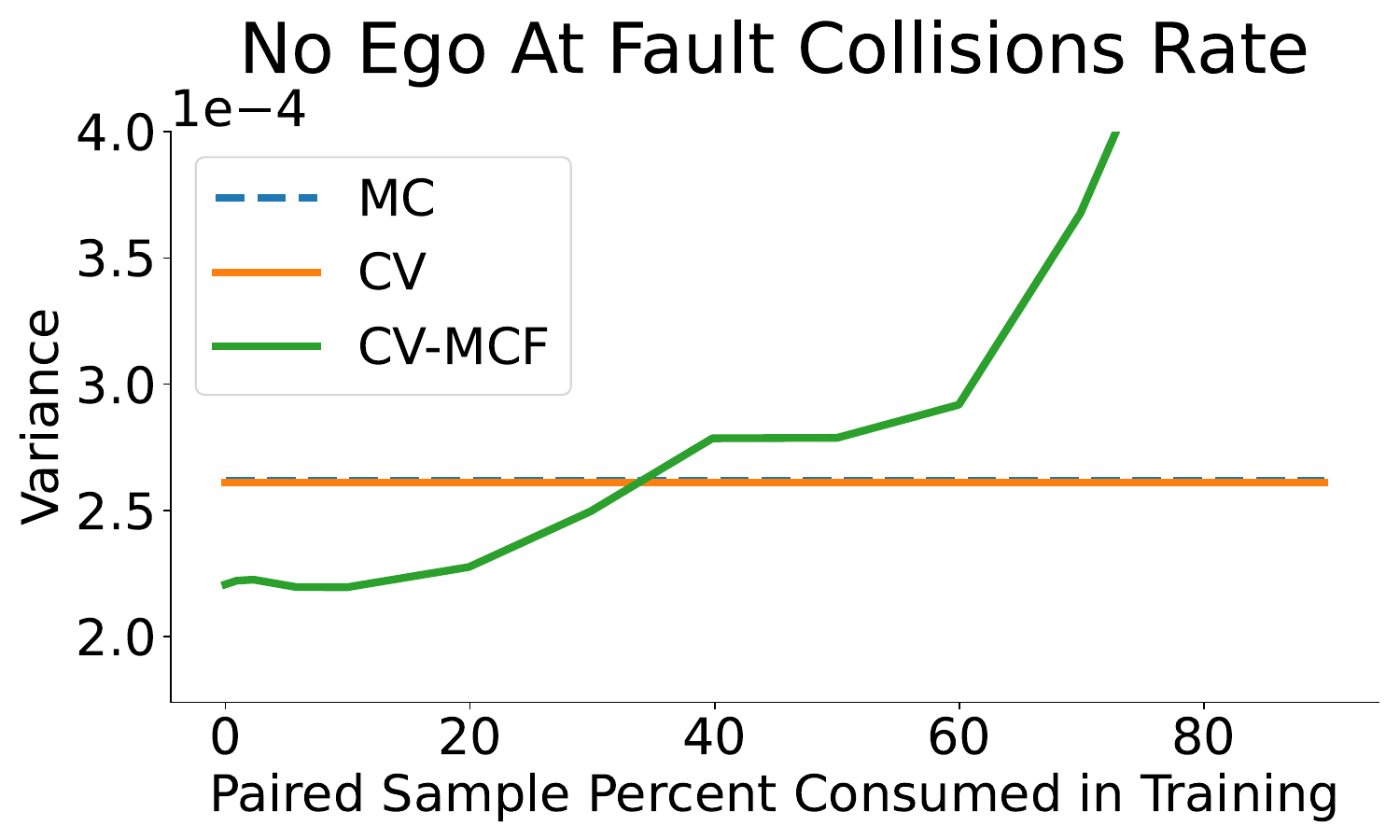}
    \caption{No ego at fault collisions metric, ``near\_multiple\_vehicles'' scenario class. MCF trained on out-of-domain data plus increasing in-domain data; $\npaired=432$, $\nsurrogate=1008$.}
    \vspace{0.1in}
    \end{subfigure}
    \hfill
    \begin{subfigure}[t]{0.49\textwidth}
    \includegraphics[width=\textwidth]{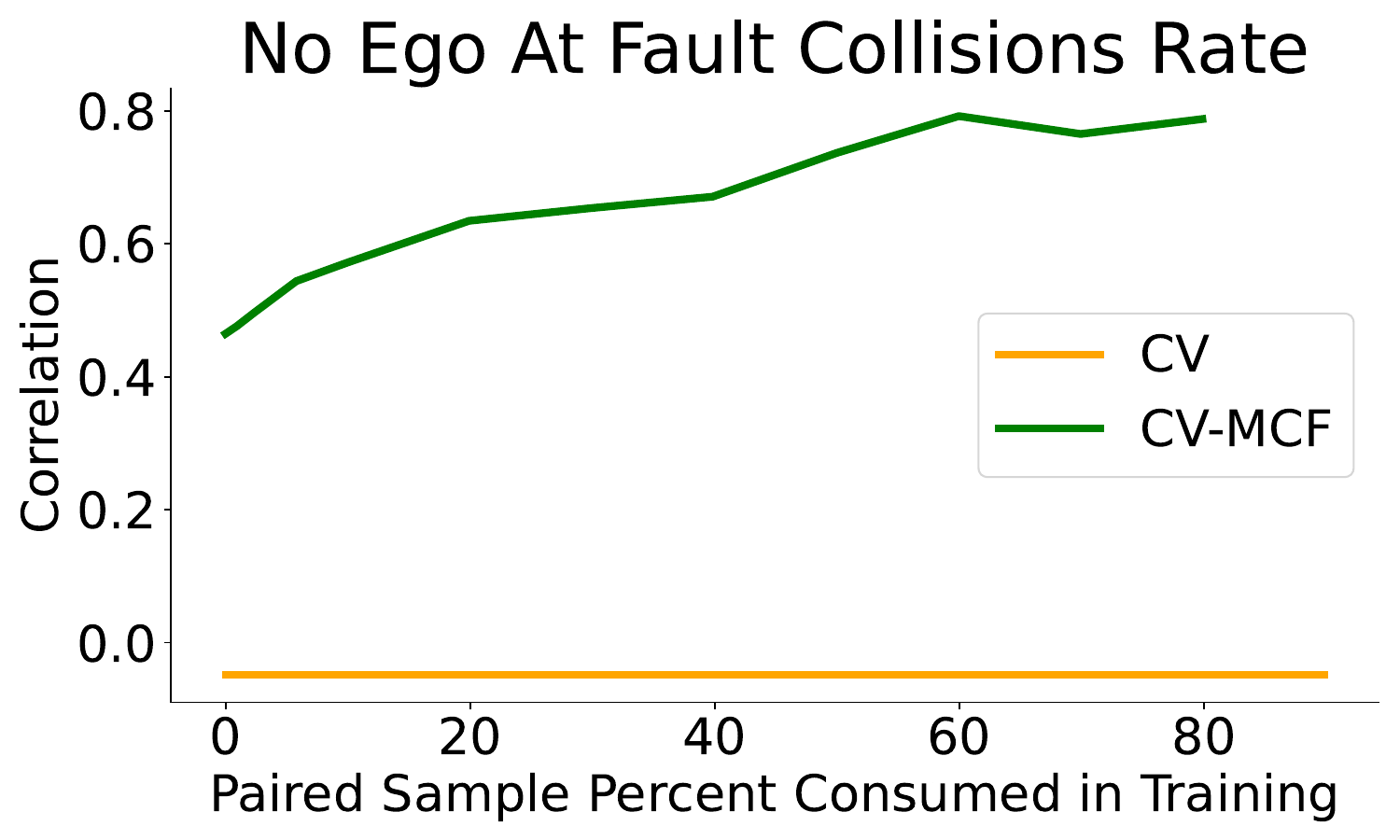}
     \caption{Corresponding correlation for the no ego at fault collisions metric, ``near\_multiple\_vehicles'' scenario class; $\npaired=432$, $\nsurrogate=1008$. }
     \vspace{0.1in}
    \end{subfigure}
    \begin{subfigure}[t]{0.49\textwidth}
        \includegraphics[width=\textwidth]{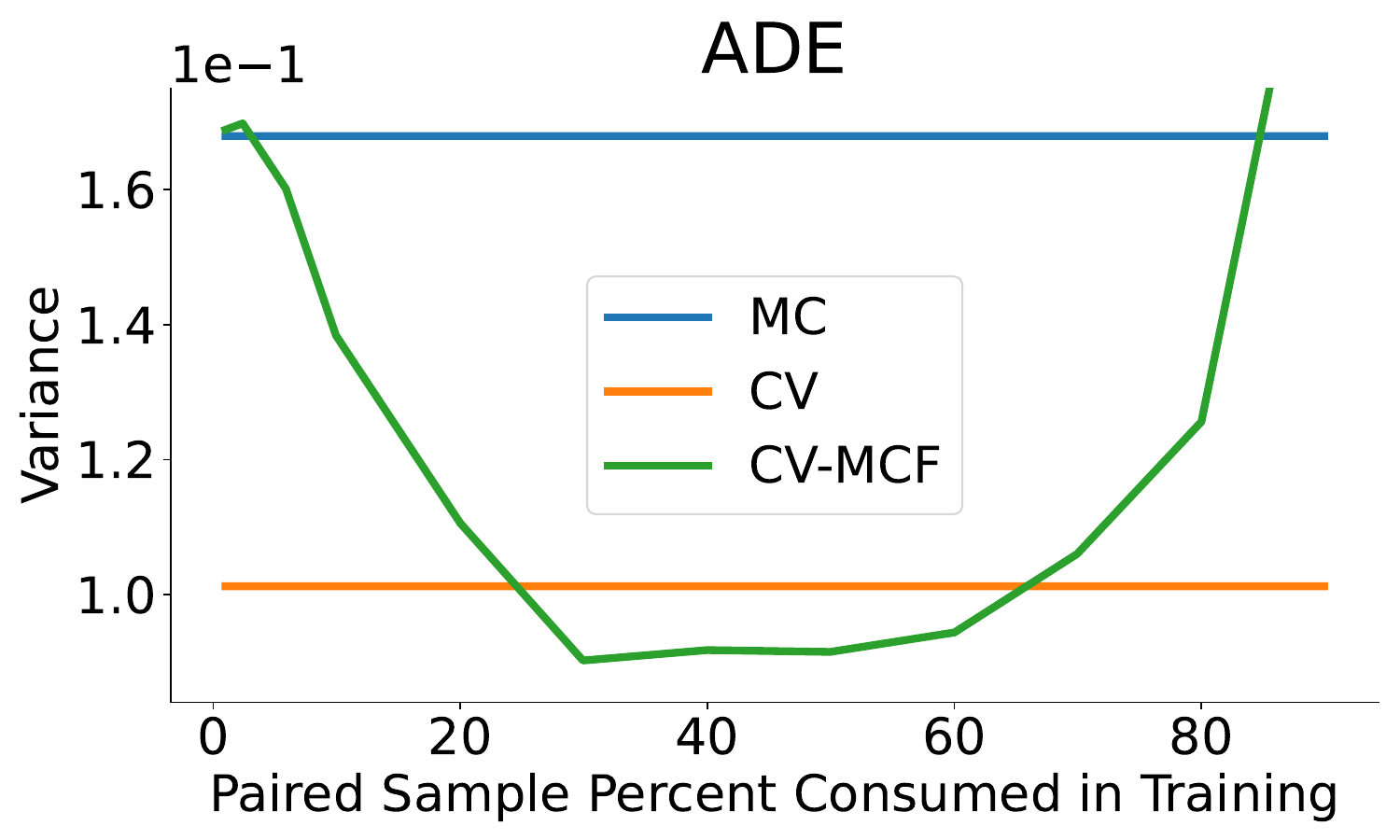}
        \caption{ADE metric, ``near\_long\_vehicle'' scenario class. MCF trained on in-domain samples only; $\npaired=715$, $\nsurrogate=1669$. }
    \end{subfigure}
    \hfill
    \begin{subfigure}[t]{0.49\textwidth}
    \includegraphics[width=\textwidth]{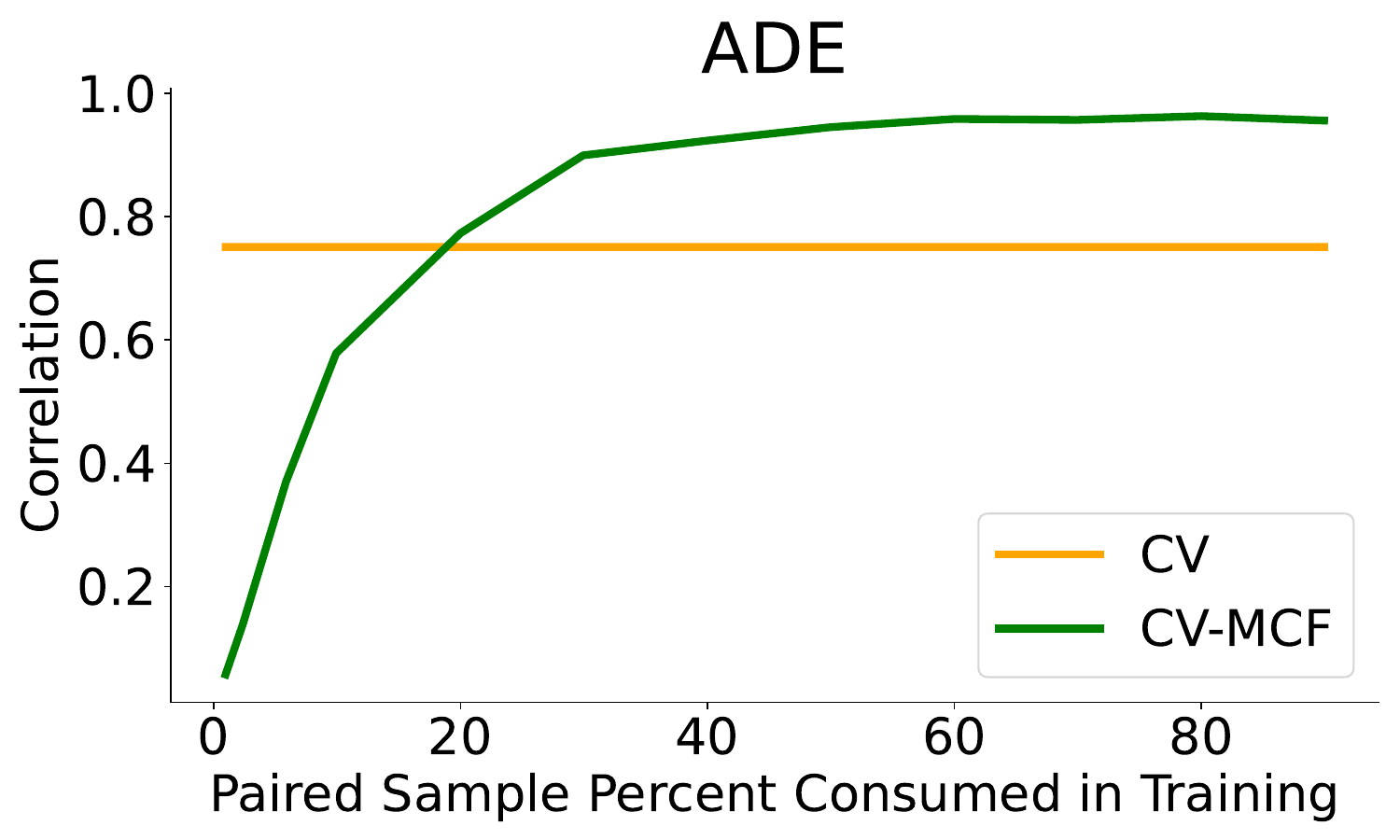}
     \caption{Corresponding correlations for the ADE metric, ``near\_long\_vehicle'' scenario class; $\npaired=715$, $\nsurrogate=1669$.}
    \end{subfigure}
    \vspace{2mm}
    \caption{Variance and corresponding correlation vs. fraction of samples used to train the MCF for different metrics, scenario classes, and training approaches. All plots averaged over 10 random trials.}
    \label{fig:nuplan_paired_sample_variance_appendix}
\end{figure}

Fig.~\ref{fig:nuplan_paired_sample_variance_appendix} plots the estimator variance as a function of the fraction of paired samples used in training the MCF (holding \(\nsurrogate\) constant).  These plots confirm that increasing correlation strength (e.g. through using more samples for training) yields lower variance only when the increased correlation outweighs the reduction in available control variates samples. Note that in panels (a) and (b), the initial correlation is low, but even a small training set yields a substantial increase in correlation. Panels (c) and (d) show initial variance reduction from improved correlation, but this benefit diminishes as the number of paired samples available for the control variate computation decreases further, and becomes insufficient to offset further correlation gains.

\subsection{Real-World Autonomous Driving Performance}

Data was collected from on-road testing of an end-to-end autonomous vehicle policy. Non-overlapping events were selected according to domain-specific definitions, including intersections, curvy roads, and stop events. Each selected event was converted into a simulation scenario by replaying 2 seconds of recorded log data, followed by 8 seconds of closed-loop simulation. Metrics such as average lane-centering error and distance to the nearest other agent at 10 seconds were computed from both the in-vehicle (real-world) and simulated logs. Due to the high correlation observed between simulated and real-world metrics, employing a metric correlator function did not provide additional benefit compared to the direct use of the control variate estimator.

\subsection{Quadruped Velocity Tracking}
Data was collected for the quadruped as follows. 
Define $F: \mathcal{X} \rightarrow \mathbb{R}$ as the real-world evaluation metric, computed through these steps:
\begin{enumerate}
 \item Initialize the quadruped at zero velocity.
 \item Deploy the reinforcement learning policy with the command velocity $X$.
 \item Allow the policy to run on the quadruped for two seconds, recording the actual velocity at each timestep as $X^{(t)}$, for $t = 1, \dots, T$, where $T = 100$ given a $50$ Hz sampling rate.
 \item Compute the average relative velocity tracking error:
 \begin{equation}
     \eta = \frac{1}{T} \sum_{t=1}^T \frac{\|X^{(t)} - X\|}{\| X \|}.
 \end{equation}
\end{enumerate}

We conducted $n=200$ paired simulation-real-world evaluations of $\eta$, supplemented by $k = 400$ simulation-only evaluations.

To train the MCF, we used a simple multilayer perceptron (MLP) that takes as input the concatenation of $X$ and the simulation metric, and outputs a prediction of the real-world metric. The MLP consists of two hidden layers, each with 4 units. To ensure that the predicted outputs are bounded between $0$ and $1$ (consistent with the empirical range of the real-world metrics), we apply a sigmoid activation function at the output layer.  

\section{Cost-Constrained Budget Allocation}

In practice, collecting different types of measurements may incur different costs. For example, real-world measurements $\metric_i$ are often expensive, while surrogate measurements $\surrogate_i$ (e.g. from simulation) are comparatively inexpensive. Given a total cost budget $C$ and per-sample costs:
\begin{itemize}
	\item $c_f$: the cost of generating metric $\metric_i$
	\item $c_g$: the cost of generating surrogate $\surrogate_i$
\end{itemize}
we can allocate the budget between $n$ paired samples and $k$ additional unpaired surrogate samples to minimize the variance of the control variate estimator: 
\begin{equation}
\label{eq:optimal_allocation}
\begin{aligned}
	&\min_{\substack{n, k \in \mathbb{Z}{\ge 0}}} \ \Var(\hat\mu_\mathrm{CV}) = \frac{\Var(F)}{n} \left( 1 - \frac{k}{n+k} \rho^2 \right) \\
	&\text{subject to: } \quad n c_f + (n+k) c_g \le C.
\end{aligned}
\end{equation}

\begin{figure}
    \centering
    \begin{subfigure}[t]{0.8\textwidth}
    \includegraphics[width=\textwidth]{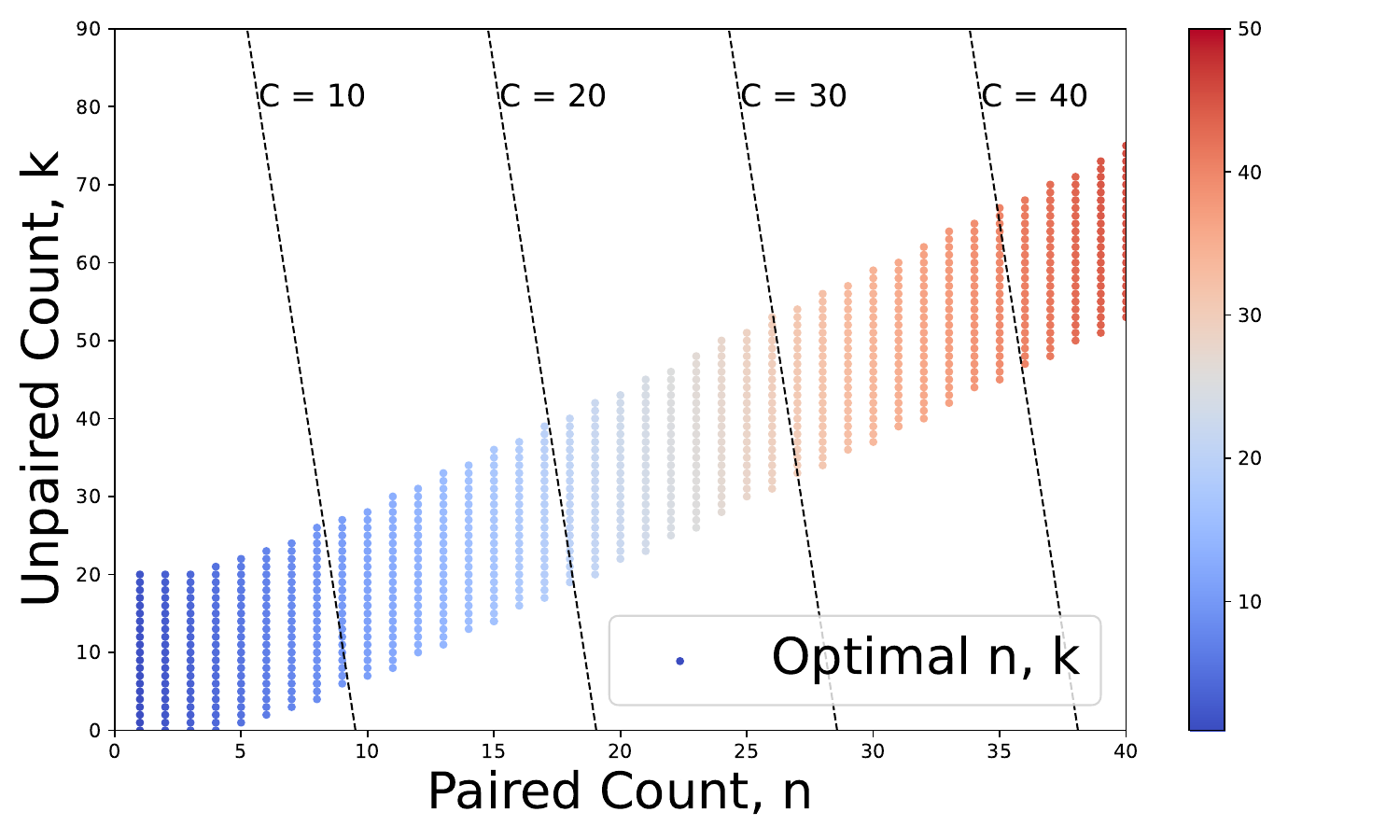}
    \caption{$c_f = 1.0, c_g = 0.05, \rho = 0.5$. }
    \end{subfigure}
     \begin{subfigure}[t]{0.8\textwidth}
    \includegraphics[width=\textwidth]{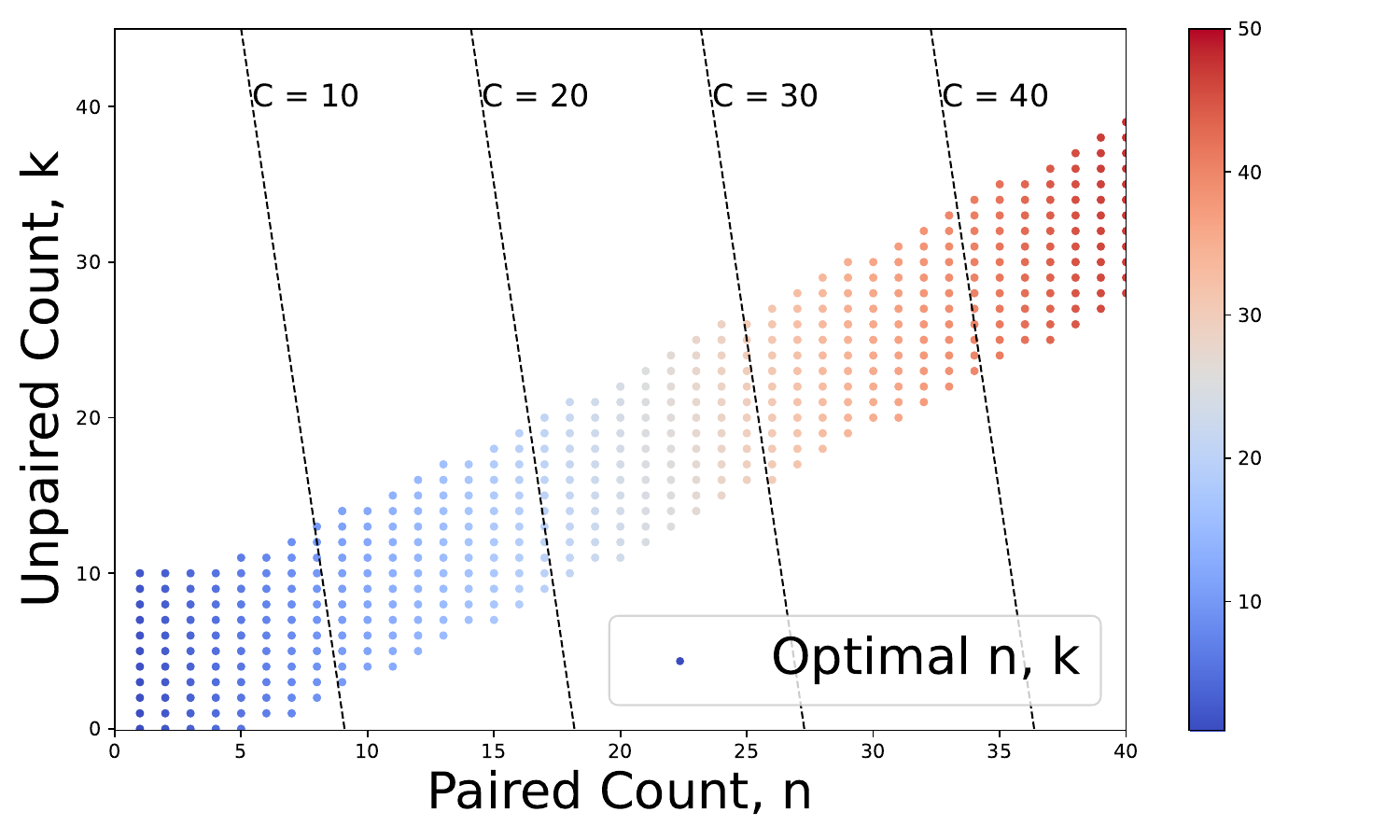}
    \caption{$c_f = 1.0, c_g = 0.1, \rho = 0.5$. }
    \end{subfigure}
    \begin{subfigure}[t]{0.8\textwidth}
    \includegraphics[width=\textwidth]{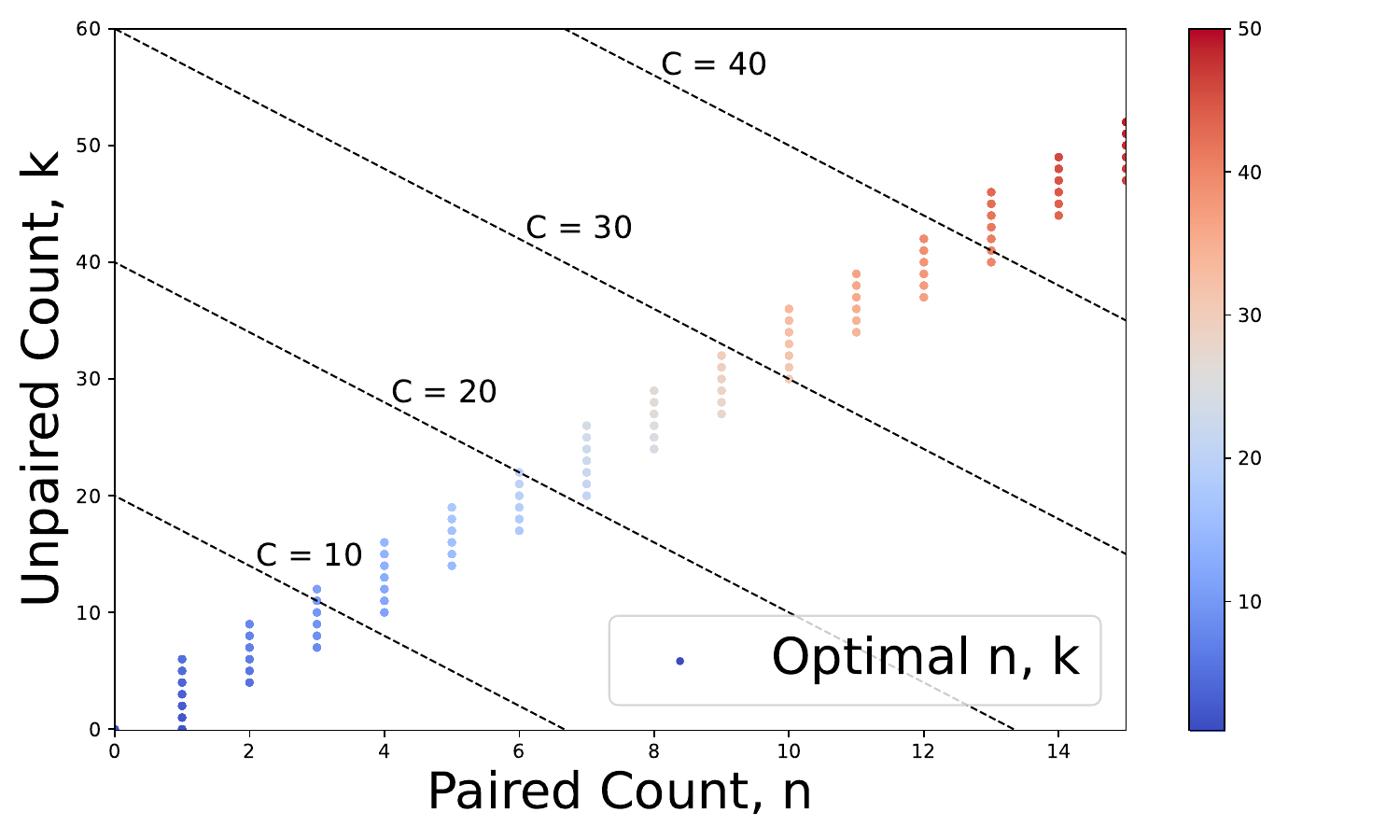}
    \caption{ $c_f = 1.0, c_g = 0.5, \rho = 0.95$.}
    \end{subfigure}
    \caption{Optimal solutions $(n, k)$ to Eq.~\ref{eq:optimal_allocation} for selected cost limits $C$. Lines of constant cost are shown at lower granularity.}
    \label{fig:optimal_allocation}
\end{figure}

Figure~\ref{fig:optimal_allocation} illustrates the optimal $(n, k)$ for various cost limits $C$. For large $C$, the optimal $n$–$k$ relationship is approximately linear.

\paragraph{Closed‐Form Solution (Continuous Relaxation).}
If we relax $n$ and $k$ to be positive real numbers, the optimal allocation can be derived in closed form. Using the cost constraint
\[
n c_f + (n + k) c_g = C
\]
and substituting $k = C/c_g - (c_f + c_g)n / c_g$, minimizing Eq. \ref{eq:optimal_allocation} yields:
\[
\begin{aligned}
    n^\star &= \frac{C\,\sqrt{1-\rho^2}}{\,c_f\sqrt{1-\rho^2} + \sqrt{c_f c_g}\,\rho\,}, \\
    k^\star &= \frac{C\bigl(\sqrt{c_f c_g}\,\rho - c_g\,\sqrt{1-\rho^2}\bigr)}
{c_g\,\bigl(\sqrt{c_f c_g}\,\rho + c_f\,\sqrt{1-\rho^2}\bigr)}\, .
\end{aligned}
\]

These values can be rounded to integers for practical use. These optimal values highlight that a higher correlation $\rho$ favors collecting more inexpensive unpaired surrogate samples, while a lower $\rho$ shifts the budget towards paired samples.

%% file: main.bbl
\begin{thebibliography}{20}
\providecommand{\natexlab}[1]{#1}
\providecommand{\url}[1]{\texttt{#1}}
\expandafter\ifx\csname urlstyle\endcsname\relax
  \providecommand{\doi}[1]{doi: #1}\else
  \providecommand{\doi}{doi: \begingroup \urlstyle{rm}\Url}\fi

\bibitem[Kalra and Paddock(2016)]{Kalra2016}
N.~Kalra and S.~M. Paddock.
\newblock Driving to safety: How many miles of driving would it take to demonstrate autonomous vehicle reliability?
\newblock \emph{Transportation Research Part A: Policy and Practice}, 94:\penalty0 182--193, 2016.
\newblock ISSN 0965-8564.
\newblock \doi{https://doi.org/10.1016/j.tra.2016.09.010}.
\newblock URL \url{https://www.sciencedirect.com/science/article/pii/S0965856416302129}.

\bibitem[Caesar et~al.(2020)Caesar, Bankiti, Lang, Vora, Liong, Xu, Krishnan, Pan, Baldan, and Beijbom]{caesar2020nuscenes}
H.~Caesar, V.~Bankiti, A.~H. Lang, S.~Vora, V.~E. Liong, Q.~Xu, A.~Krishnan, Y.~Pan, G.~Baldan, and O.~Beijbom.
\newblock nuscenes: A multimodal dataset for autonomous driving.
\newblock In \emph{Proceedings of the IEEE/CVF conference on computer vision and pattern recognition}, pages 11621--11631, 2020.

\bibitem[Karnchanachari et~al.(2024)Karnchanachari, Geromichalos, Tan, Li, Eriksen, Yaghoubi, Mehdipour, Bernasconi, Fong, Guo, et~al.]{karnchanachari2024towards}
N.~Karnchanachari, D.~Geromichalos, K.~S. Tan, N.~Li, C.~Eriksen, S.~Yaghoubi, N.~Mehdipour, G.~Bernasconi, W.~K. Fong, Y.~Guo, et~al.
\newblock Towards learning-based planning: The nuplan benchmark for real-world autonomous driving.
\newblock In \emph{2024 IEEE International Conference on Robotics and Automation (ICRA)}, pages 629--636. IEEE, 2024.

\bibitem[Collaboration et~al.(2023)Collaboration, O'Neill, Rehman, Gupta, Maddukuri, Gupta, Padalkar, Lee, Pooley, Gupta, Mandlekar, Jain, Tung, Bewley, Herzog, Irpan, Khazatsky, Rai, Gupta, Wang, Kolobov, Singh, Garg, Kembhavi, Xie, Brohan, Raffin, Sharma, Yavary, Jain, Balakrishna, Wahid, Burgess-Limerick, Kim, Schölkopf, Wulfe, Ichter, Lu, Xu, Le, Finn, Wang, Xu, Chi, Huang, Chan, Agia, Pan, Fu, Devin, Xu, Morton, Driess, Chen, Pathak, Shah, Büchler, Jayaraman, Kalashnikov, Sadigh, Johns, Foster, Liu, Ceola, Xia, Zhao, Frujeri, Stulp, Zhou, Sukhatme, Salhotra, Yan, Feng, Schiavi, Berseth, Kahn, Yang, Wang, Su, Fang, Shi, Bao, Amor, Christensen, Furuta, Bharadhwaj, Walke, Fang, Ha, Mordatch, Radosavovic, Leal, Liang, Abou-Chakra, Kim, Drake, Peters, Schneider, Hsu, Vakil, Bohg, Bingham, Wu, Gao, Hu, Wu, Wu, Sun, Luo, Gu, Tan, Oh, Wu, Lu, Yang, Malik, Silvério, Hejna, Booher, Tompson, Yang, Salvador, Lim, Han, Wang, Rao, Pertsch, Hausman, Go, Gopalakrishnan, Goldberg, Byrne, Oslund, Kawaharazuka, Black,
  Lin, Zhang, Ehsani, Lekkala, Ellis, Rana, Srinivasan, Fang, Singh, Zeng, Hatch, Hsu, Itti, Chen, Pinto, Fei-Fei, Tan, Fan, Ott, Lee, Weihs, Chen, Lepert, Memmel, Tomizuka, Itkina, Castro, Spero, Du, Ahn, Yip, Zhang, Ding, Heo, Srirama, Sharma, Kim, Kanazawa, Hansen, Heess, Joshi, Suenderhauf, Liu, Palo, Shafiullah, Mees, Kroemer, Bastani, Sanketi, Miller, Yin, Wohlhart, Xu, Fagan, Mitrano, Sermanet, Abbeel, Sundaresan, Chen, Vuong, Rafailov, Tian, Doshi, Martin-Martin, Baijal, Scalise, Hendrix, Lin, Qian, Zhang, Mendonca, Shah, Hoque, Julian, Bustamante, Kirmani, Levine, Lin, Moore, Bahl, Dass, Sonawani, Tulsiani, Song, Xu, Haldar, Karamcheti, Adebola, Guist, Nasiriany, Schaal, Welker, Tian, Ramamoorthy, Dasari, Belkhale, Park, Nair, Mirchandani, Osa, Gupta, Harada, Matsushima, Xiao, Kollar, Yu, Ding, Davchev, Zhao, Armstrong, Darrell, Chung, Jain, Kumar, Vanhoucke, Zhan, Zhou, Burgard, Chen, Chen, Wang, Zhu, Geng, Liu, Liangwei, Li, Pang, Lu, Ma, Kim, Chebotar, Zhou, Zhu, Wu, Xu, Wang, Bisk, Dou, Cho, Lee,
  Cui, Cao, Wu, Tang, Zhu, Zhang, Jiang, Li, Li, Iwasawa, Matsuo, Ma, Xu, Cui, Zhang, Fu, and Lin]{open_x_embodiment_rt_x_2023}
O.~X.-E. Collaboration, A.~O'Neill, A.~Rehman, A.~Gupta, A.~Maddukuri, A.~Gupta, A.~Padalkar, A.~Lee, A.~Pooley, A.~Gupta, A.~Mandlekar, A.~Jain, A.~Tung, A.~Bewley, A.~Herzog, A.~Irpan, A.~Khazatsky, A.~Rai, A.~Gupta, A.~Wang, A.~Kolobov, A.~Singh, A.~Garg, A.~Kembhavi, A.~Xie, A.~Brohan, A.~Raffin, A.~Sharma, A.~Yavary, A.~Jain, A.~Balakrishna, A.~Wahid, B.~Burgess-Limerick, B.~Kim, B.~Schölkopf, B.~Wulfe, B.~Ichter, C.~Lu, C.~Xu, C.~Le, C.~Finn, C.~Wang, C.~Xu, C.~Chi, C.~Huang, C.~Chan, C.~Agia, C.~Pan, C.~Fu, C.~Devin, D.~Xu, D.~Morton, D.~Driess, D.~Chen, D.~Pathak, D.~Shah, D.~Büchler, D.~Jayaraman, D.~Kalashnikov, D.~Sadigh, E.~Johns, E.~Foster, F.~Liu, F.~Ceola, F.~Xia, F.~Zhao, F.~V. Frujeri, F.~Stulp, G.~Zhou, G.~S. Sukhatme, G.~Salhotra, G.~Yan, G.~Feng, G.~Schiavi, G.~Berseth, G.~Kahn, G.~Yang, G.~Wang, H.~Su, H.-S. Fang, H.~Shi, H.~Bao, H.~B. Amor, H.~I. Christensen, H.~Furuta, H.~Bharadhwaj, H.~Walke, H.~Fang, H.~Ha, I.~Mordatch, I.~Radosavovic, I.~Leal, J.~Liang, J.~Abou-Chakra, J.~Kim,
  J.~Drake, J.~Peters, J.~Schneider, J.~Hsu, J.~Vakil, J.~Bohg, J.~Bingham, J.~Wu, J.~Gao, J.~Hu, J.~Wu, J.~Wu, J.~Sun, J.~Luo, J.~Gu, J.~Tan, J.~Oh, J.~Wu, J.~Lu, J.~Yang, J.~Malik, J.~Silvério, J.~Hejna, J.~Booher, J.~Tompson, J.~Yang, J.~Salvador, J.~J. Lim, J.~Han, K.~Wang, K.~Rao, K.~Pertsch, K.~Hausman, K.~Go, K.~Gopalakrishnan, K.~Goldberg, K.~Byrne, K.~Oslund, K.~Kawaharazuka, K.~Black, K.~Lin, K.~Zhang, K.~Ehsani, K.~Lekkala, K.~Ellis, K.~Rana, K.~Srinivasan, K.~Fang, K.~P. Singh, K.-H. Zeng, K.~Hatch, K.~Hsu, L.~Itti, L.~Y. Chen, L.~Pinto, L.~Fei-Fei, L.~Tan, L.~J. Fan, L.~Ott, L.~Lee, L.~Weihs, M.~Chen, M.~Lepert, M.~Memmel, M.~Tomizuka, M.~Itkina, M.~G. Castro, M.~Spero, M.~Du, M.~Ahn, M.~C. Yip, M.~Zhang, M.~Ding, M.~Heo, M.~K. Srirama, M.~Sharma, M.~J. Kim, N.~Kanazawa, N.~Hansen, N.~Heess, N.~J. Joshi, N.~Suenderhauf, N.~Liu, N.~D. Palo, N.~M.~M. Shafiullah, O.~Mees, O.~Kroemer, O.~Bastani, P.~R. Sanketi, P.~T. Miller, P.~Yin, P.~Wohlhart, P.~Xu, P.~D. Fagan, P.~Mitrano, P.~Sermanet,
  P.~Abbeel, P.~Sundaresan, Q.~Chen, Q.~Vuong, R.~Rafailov, R.~Tian, R.~Doshi, R.~Martin-Martin, R.~Baijal, R.~Scalise, R.~Hendrix, R.~Lin, R.~Qian, R.~Zhang, R.~Mendonca, R.~Shah, R.~Hoque, R.~Julian, S.~Bustamante, S.~Kirmani, S.~Levine, S.~Lin, S.~Moore, S.~Bahl, S.~Dass, S.~Sonawani, S.~Tulsiani, S.~Song, S.~Xu, S.~Haldar, S.~Karamcheti, S.~Adebola, S.~Guist, S.~Nasiriany, S.~Schaal, S.~Welker, S.~Tian, S.~Ramamoorthy, S.~Dasari, S.~Belkhale, S.~Park, S.~Nair, S.~Mirchandani, T.~Osa, T.~Gupta, T.~Harada, T.~Matsushima, T.~Xiao, T.~Kollar, T.~Yu, T.~Ding, T.~Davchev, T.~Z. Zhao, T.~Armstrong, T.~Darrell, T.~Chung, V.~Jain, V.~Kumar, V.~Vanhoucke, W.~Zhan, W.~Zhou, W.~Burgard, X.~Chen, X.~Chen, X.~Wang, X.~Zhu, X.~Geng, X.~Liu, X.~Liangwei, X.~Li, Y.~Pang, Y.~Lu, Y.~J. Ma, Y.~Kim, Y.~Chebotar, Y.~Zhou, Y.~Zhu, Y.~Wu, Y.~Xu, Y.~Wang, Y.~Bisk, Y.~Dou, Y.~Cho, Y.~Lee, Y.~Cui, Y.~Cao, Y.-H. Wu, Y.~Tang, Y.~Zhu, Y.~Zhang, Y.~Jiang, Y.~Li, Y.~Li, Y.~Iwasawa, Y.~Matsuo, Z.~Ma, Z.~Xu, Z.~J. Cui, Z.~Zhang, Z.~Fu,
  and Z.~Lin.
\newblock Open {X-E}mbodiment: Robotic learning datasets and {RT-X} models.
\newblock \url{https://arxiv.org/abs/2310.08864}, 2023.

\bibitem[Gulino et~al.(2023)Gulino, Fu, Luo, Tucker, Bronstein, Lu, Harb, Pan, Wang, Chen, et~al.]{gulino2023waymax}
C.~Gulino, J.~Fu, W.~Luo, G.~Tucker, E.~Bronstein, Y.~Lu, J.~Harb, X.~Pan, Y.~Wang, X.~Chen, et~al.
\newblock Waymax: An accelerated, data-driven simulator for large-scale autonomous driving research.
\newblock \emph{Advances in Neural Information Processing Systems}, 36:\penalty0 7730--7742, 2023.

\bibitem[Urtasan(2025)]{urtasan2025simulator}
R.~Urtasan.
\newblock Simulator realism: The new safety standard for the av industry, 2025.
\newblock URL \url{https://waabi.ai/simulator-realism-the-new-safety-standard-for-the-av-industry/}.

\bibitem[wu2()]{wu2025alpamayo}
Reference for Simulator, author and organization redacted for review.

\bibitem[Levine et~al.(2020)Levine, Kumar, Tucker, and Fu]{levine2020offline}
S.~Levine, A.~Kumar, G.~Tucker, and J.~Fu.
\newblock Offline reinforcement learning: Tutorial, review, and perspectives on open problems.
\newblock \emph{arXiv preprint arXiv:2005.01643}, 2020.

\bibitem[Precup et~al.(2000)Precup, Sutton, and Singh]{precup2000eligibility}
D.~Precup, R.~S. Sutton, and S.~Singh.
\newblock Eligibility traces for off-policy policy evaluation.
\newblock In \emph{ICML}, volume 2000, pages 759--766. Citeseer, 2000.

\bibitem[Jiang and Li(2016)]{jiang2016doubly}
N.~Jiang and L.~Li.
\newblock Doubly robust off-policy value evaluation for reinforcement learning.
\newblock In \emph{International conference on machine learning}, pages 652--661. PMLR, 2016.

\bibitem[Thomas and Brunskill(2016)]{thomas2016data}
P.~Thomas and E.~Brunskill.
\newblock Data-efficient off-policy policy evaluation for reinforcement learning.
\newblock In \emph{International conference on machine learning}, pages 2139--2148. PMLR, 2016.

\bibitem[Owen(2013)]{owen2013monte}
A.~B. Owen.
\newblock \emph{Monte Carlo theory, methods and examples}.
\newblock \url{https://artowen.su.domains/mc/}, 2013.

\bibitem[Angelopoulos et~al.(2023{\natexlab{a}})Angelopoulos, Bates, Fannjiang, Jordan, and Zrnic]{angelopoulos2023prediction}
A.~N. Angelopoulos, S.~Bates, C.~Fannjiang, M.~I. Jordan, and T.~Zrnic.
\newblock Prediction-powered inference.
\newblock \emph{Science}, 382\penalty0 (6671):\penalty0 669--674, 2023{\natexlab{a}}.

\bibitem[Angelopoulos et~al.(2023{\natexlab{b}})Angelopoulos, Duchi, and Zrnic]{angelopoulos2023ppi++}
A.~N. Angelopoulos, J.~C. Duchi, and T.~Zrnic.
\newblock Ppi++: Efficient prediction-powered inference.
\newblock \emph{arXiv preprint arXiv:2311.01453}, 2023{\natexlab{b}}.

\bibitem[Zhou et~al.(2025)Zhou, Song, and Zanette]{zhou2025accelerating}
Z.~Zhou, Y.~Song, and A.~Zanette.
\newblock Accelerating unbiased llm evaluation via synthetic feedback.
\newblock \emph{arXiv preprint arXiv:2502.10563}, 2025.

\bibitem[Boyeau et~al.(2024)Boyeau, Angelopoulos, Yosef, Malik, and Jordan]{boyeau2024autoeval}
P.~Boyeau, A.~N. Angelopoulos, N.~Yosef, J.~Malik, and M.~I. Jordan.
\newblock Autoeval done right: Using synthetic data for model evaluation.
\newblock \emph{arXiv preprint arXiv:2403.07008}, 2024.

\bibitem[Dauner et~al.(2024)Dauner, Hallgarten, Li, Weng, Huang, Yang, Li, Gilitschenski, Ivanovic, Pavone, et~al.]{dauner2024navsim}
D.~Dauner, M.~Hallgarten, T.~Li, X.~Weng, Z.~Huang, Z.~Yang, H.~Li, I.~Gilitschenski, B.~Ivanovic, M.~Pavone, et~al.
\newblock Navsim: Data-driven non-reactive autonomous vehicle simulation and benchmarking.
\newblock \emph{Advances in Neural Information Processing Systems}, 37:\penalty0 28706--28719, 2024.

\bibitem[H.~Caesar(2021)]{nuplan}
K.~T. e.~a. H.~Caesar, J.~Kabzan.
\newblock Nuplan: A closed-loop ml-based planning benchmark for autonomous vehicles.
\newblock In \emph{CVPR ADP3 workshop}, 2021.

\bibitem[Treiber et~al.(2000)Treiber, Hennecke, and Helbing]{Treiber2000CongestedTS}
M.~Treiber, A.~Hennecke, and D.~Helbing.
\newblock Congested traffic states in empirical observations and microscopic simulations.
\newblock \emph{Physical review. E, Statistical physics, plasmas, fluids, and related interdisciplinary topics}, 62 2 Pt A:\penalty0 1805--24, 2000.
\newblock URL \url{https://api.semanticscholar.org/CorpusID:1100293}.

\bibitem[Ding et~al.(2025)Ding, Veer, Leung, Cao, and Pavone]{ding2025surprise}
W.~Ding, S.~Veer, K.~Leung, Y.~Cao, and M.~Pavone.
\newblock Surprise potential as a measure of interactivity in driving scenarios.
\newblock \emph{arXiv preprint arXiv:2502.05677}, 2025.

\end{thebibliography}
